\documentclass[letterpaper,journal]{IEEEtran}
\usepackage{amsmath,amsfonts}
\usepackage{array}
\usepackage{textcomp}
\usepackage{stfloats}
\usepackage{url}
\usepackage{verbatim}
\usepackage{graphicx}

\usepackage{booktabs}
\usepackage[table]{xcolor}
\usepackage{mathtools} 
\usepackage{dsfont} 
\usepackage{algorithm}
\usepackage[noend]{algpseudocode}
\usepackage{multirow}
\usepackage{adjustbox}
\usepackage{makecell}
\usepackage{amsthm}
\usepackage{circledsteps}
\usepackage{wrapfig} 
\usepackage{subcaption}
\usepackage{orcidlink}

\newtheorem{proposition}{Proposition}
\newtheorem{lemma}{Lemma}
\newtheorem{theorem}{Theorem}

\newcommand{\major}[1]{\textcolor{black}{#1}}

\begin{document}

\title{FedQUIT: On-Device Federated Unlearning \\
via a Quasi-Competent Virtual Teacher}

\author{~Alessio~Mora,
        ~Lorenzo~Valerio,
        ~Paolo~Bellavista,
        ~Andrea~Passarella
\thanks{Alessio Mora and Paolo Bellavista are with the Department of Computer Science and Engineering (DISI), University of Bologna, Bologna, Italy (e-mail: alessio.mora@unibo.it, paolo.bellavista@unibo.it).\\
Lorenzo Valerio and Andrea Passarella are with the Institute of Informatics and Telematics - National Research Council of Italy, Pisa, Italy (IIT-CNR) (e-mail: lorenzo.valerio@iit.cnr.it, a.passarella@iit.cnr.it).

}%
}

\markboth{Journal of \LaTeX\ Class Files,~Vol.~14, No.~8, August~2021}%
{Shell \MakeLowercase{\textit{et al.}}: A Sample Article Using IEEEtran.cls for IEEE Journals}


\maketitle
\begin{abstract}
Federated Learning (FL) enables the collaborative training of machine learning models without requiring centralized collection of user data. To comply with the right to be forgotten, FL clients should be able to request the removal of their data contributions from the global model. In this paper, we propose FedQUIT, a novel unlearning algorithm that operates directly on client devices that request to remove its contribution. Our method leverages knowledge distillation to remove the influence of the target client’s data from the global model while preserving its generalization ability. FedQUIT adopts a teacher–student framework, where a modified version of the current global model serves as a virtual teacher and the client's model acts as the student. The virtual teacher is obtained by manipulating the global model’s outputs on forget data, penalizing the confidence assigned to the true class while preserving relationships among outputs of non-true classes, to simultaneously induce forgetting and retain useful knowledge. As a result, FedQUIT achieves unlearning without making any additional assumption over the standard FedAvg protocol. Evaluation across diverse datasets, data heterogeneity levels, and model architectures shows that FedQUIT achieves superior or comparable unlearning efficacy compared to six state-of-the-art methods, while significantly reducing cumulative communication and computational overhead relative to retraining from scratch.
\end{abstract}


\section{Introduction}
\label{sec:intro}


Federated Learning (FL) trains a shared global model by periodically aggregating ephemeral model updates that are locally computed by users' devices on private data \cite{mcmahan2017communication}, avoiding the transfer and collection of unprocessed data. However, privacy regulations such as the GDPR \cite{GDPR2016a} also require the enforcement of the right to be forgotten, giving users the ability to request the deletion of their personal data upon withdrawal of usage consent. As widely demonstrated in prior work \cite{shokri2017membership, song2019privacy, song2021systematic}, deep learning models can memorize and leak sensitive information from training data. Consequently, simply deleting the data samples to forget is insufficient, both in centralized \cite{golatkar2020eternal, xu2023machine} and in federated settings \cite{romandini2024federated, liu2024survey}. To address the latter challenge, a growing body of federated unlearning (FU) mechanisms has emerged. Early FU methods perform unlearning by leveraging stored historical model updates to estimate and subtract the contribution of the data to forget, and then recalibrate the global model accordingly \cite{federaser}, often combined with rapid retraining strategies \cite{liu2022right, wu2022unlearningdistillation, zhang2025model}. More recent approaches avoid retaining historical updates, since this poses storage and privacy risks, and typically adopt a two-stage design: an unlearning phase to remove the influence of the forget data, followed by a recovery phase that restores utility via training on the retained data. Nevertheless, these methods still present drawbacks: some require extensive hyperparameter tuning and can overwrite knowledge to retain, causing model erasure \cite{halimi2022federated}; others use multi-round unlearning, keeping the requesting client actively involved and unable to leave the federation \cite{zhao2023federated, pan2025federated}, increasing coordination; some introduce auxiliary unlearning modules trained during learning \cite{gu2024unlearning}, forcing changes to local training and model architecture; others depend on a priori identification of critical layers in the model \cite{zhong2025unlearning}; or lack selectivity in removing client-specific contributions \cite{khalil2025not}. Furthermore, as we demonstrate in our empirical evaluations, these methods often fail to be both effective in forgetting and efficient in restoring model utility (i.e., with minimal communication and computation overhead), particularly when data are homogeneously distributed.


\begin{figure}[t!]
    \centering
    \includegraphics[width=0.95\columnwidth]{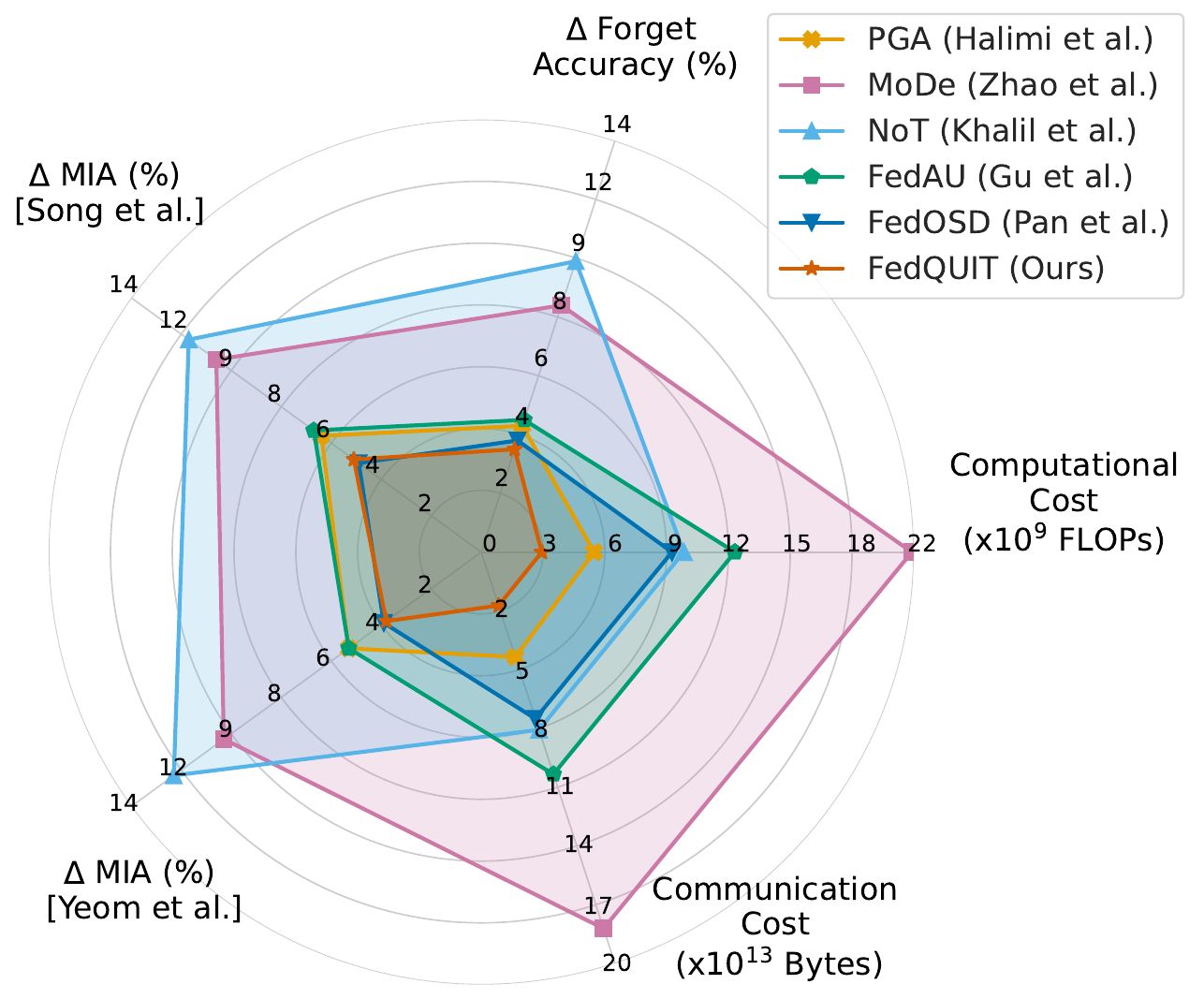}
    \caption{\textbf{FedQUIT vs. SOTA on CIFAR-100} (non-IID, ResNet-18; 10 clients, one unlearning request). Radar shows post-recovery absolute gap to \ \textit{Retrain} for accuracy and MIAs on forget data, and the required computational/communication overhead. Smaller polygons indicate better unlearning; see Sec.~\ref{sec:exp_design} for experimental design.}

    \label{fig:radar_intro}
\end{figure}

In this paper, we propose FedQUIT, a novel mechanism for client and sample unlearning that does not rely on historical updates or auxiliary data and requires a single on-device unlearning round while the target client is still connected (before leaving). 
During the unlearning phase, the model is trained to mimic the output of a \emph{virtual teacher} on the local forget data. Concretely, our virtual teacher mirrors the output of the current global model on the forget data while selectively penalizing the true-class score and preserving the relationships among the non-true classes. Since high-confidence predictions are indicative of memorization \cite{yeom2018privacy, song2019privacy, song2021systematic, yeleave}, we explicitly reduce confidence in the correct label to induce forgetting, while preserving the inter-class structure among the non-true classes \cite{hinton2015distilling, phuong2019towards, lee2021preservation} to preserve model utility. Under standard smoothness/bounded-gradient assumptions, this distillation update provides a controlled “forgetting signal” and induces only a bounded parameter perturbation, which in turn implies that resuming standard FedAvg from the unlearned snapshot retains the usual convergence rate up to an initialization-dependent offset (i.e., a quantifiable “price of forgetting”).

As shown in Figure \ref{fig:radar_intro} (full results in Table \ref{table:full_results}), among state-of-the-art (SOTA) FU methods, FedQUIT consistently achieves performance closest to the gold-standard retraining baseline across multiple accuracy metrics, while significantly reducing cumulative communication and computation costs.

\textbf{Contributions.}
The contributions of the paper are summarized as follows:
\begin{itemize}
    \item We introduce FedQUIT, an efficient on-device FU method that induces forgetting via lightweight knowledge distillation (KD) from a quasi-competent virtual teacher that lowers the-true class score while preserving non-true class geometry.
    \item We provide theoretical insights showing that distilling to our virtual teacher (preserving the non-true geometry and penalizing only the true-class output) (i) provides a controlled forgetting signal; and (ii) induces a bounded parameter shift, so that FedAvg resumed from the unlearned model retains its standard convergence guarantees.
    \item We conduct evaluations across four datasets, both IID and non-IID data distributions, and three model architectures, comparing FedQUIT to six SOTA baselines. FedQUIT achieves superior unlearning: it better or comparably approximates the gold-standard retrained model in forget metrics, while significantly reducing cumulative communication and computational cost.
\end{itemize} 
Our code is available at: \url{https://anonymous.4open.science/r/FedQUIT}.

\section{Related Work}
\label{related_work}

We group existing FU methods according to their core unlearning mechanism. For each category, we discuss the underlying design choices and their practical limitations, and highlight how \textsc{FedQUIT} departs from or advances beyond prior approaches when relevant. We conclude this section with a dedicated discussion on the tuning complexity and hyperparameter sensitivity of existing methods, contrasting them with \textsc{FedQUIT}.

\noindent \textbf{History-based subtraction.} 
These methods retain per–client update histories and reconstruct a sanitized model by removing the target client’s contribution. FedEraser \cite{federaser} recalibrates historical updates excluding the target contribution; the work in \cite{liu2022right} accelerates retraining using quasi Newton updates. However, methods that remove a client by reconstructing or compensating its past contributions require storing server-side histories of per-client updates. We note that: (1) Linking per-client update histories to specific clients requesting unlearning may weaken the FL's privacy design; (2) to some extent it violates the \emph{ephemeral updates} requirement of the FL framework \cite{kairouz2019advances}; and (3) at scale, considering a massive number of clients \cite{dean2012large}, several FL rounds, and multiple simultaneous learning tasks, the corresponding storage requirements might become impractical.

\noindent \textbf{History-based subtraction and KD recovery on public data.} Other works subtract the target’s updates and then use KD on public proxy data to recover utility \cite{wu2022unlearningdistillation, wu2023unlearning, zhang2025model}. However, to be useful, this public dataset should be semantically similar (and balanced between classes) to the data in the federation \cite{nayak2021effectiveness}; assuming the existence and leveraging the usage of a semantically similar public dataset in FL has been previously argued to be unrealistic \cite{mora2022knowledge}.

Unlike prior KD-based FU methods that rely on stored historical updates and distill the pre-unlearning global model using external or public data, FedQUIT introduces a different distillation mechanism. While history-based approaches first roll back the target client’s updates and then use regular KD \cite{hinton2015distilling} on a proxy dataset to recover utility, FedQUIT uses no stored updates and no auxiliary or public data. It constructs a virtual teacher directly from the current global model and performs distillation on the requesting client’s own data enabling a fully on-device and single-round unlearning step.

\noindent \textbf{Gradient manipulation.}
These methods manipulate gradients computed on the forget data before applying them to the unlearned model.
PGA \cite{halimi2022federated} performs projected gradient ascent on forget data, enforcing an $l2$-norm constraint around a reference model. However, it requires maintaining client-side state such as the last update to build their reference model and for projections and clipping. FedOSD \cite{pan2025federated} introduces a two-stage, multi-round method: first, it enforces forgetting by training the target client with a custom Cross-Entropy and updating the server via an orthogonal steepest-descent direction; then, it post-trains while projecting retained gradients to prevent conflicts. However, procedures that require multiple rounds keep the requesting client active for an extended period (all the unlearning rounds), which may be unpractical, and increase coordination cost. 

\noindent \textbf{Direct weight manipulation.} FUSED \cite{zhong2025unlearning} preemptively identifies critical layers and trains sparse unlearning adapters only on retained clients. NoT \cite{khalil2025not} negates the weights of the first layer, producing the same unlearned model regardless of forget data. This lack of selectivity leaves residual influence and often prolongs recovery, as shown in our experimental results.

\noindent \textbf{Supervision perturbation.}
These methods modify the supervision signal on forget data.
FedUNRAN \cite{mora2024fedunran} fine tunes the global model on local forget data with random labels, in a single-round unlearning phase. FedAU \cite{gu2024unlearning} uses random labels on forget data to learn an auxiliary module, and then combines this component with the original model to produce the unelarned model.
We note that random-label perturbations can be viewed as supervision via a teacher model that outputs random
one-hot labels (see Appendix A, subsection A). As a result, it falls outside the geometry-preserving teacher class assumed in our theoretical framework (see Section \ref{sec:theory}).

FedQUIT (our method) belongs to this supervision-perturbation family, but preserves the non–true-class
structure of the current global model (true-class logit penalized, other logits preserved), thereby
ensuring a bounded perturbation.
MoDe \cite{zhao2023federated} proposes a multi-round mechanism that alternates, in each round, degradation and memory guidance using hard labels from a degraded model. Similar to FedOSD \cite{pan2025federated}, this multi-round design requires the requesting client to remain active in the federation for several rounds before leaving. In contrast, FedQUIT performs unlearning in a single round.

\noindent \textbf{Tuning and hyperparameters.}
Many FU methods introduce additional schedules, projections, or thresholds beyond standard FL training, resulting in substantial hyperparameter tuning requirements. 
For instance, FedEraser~\cite{federaser} depends critically on (i) the retention interval for stored updates and (ii) the number of calibration epochs, whose interaction strongly affects performance. 
FUKD~\cite{wu2022unlearningdistillation} requires selecting the size and composition of the proxy dataset used for distillation, as well as tuning the learning rate, number of epochs, and batch size for KD, all of which depend on the chosen transfer dataset and model.
PGA~\cite{halimi2022federated} involves multiple sensitive hyperparameters, including a distance-based stopping threshold, learning rate, number of local ascent epochs, large batch sizes, and gradient clipping norms to ensure stability.
Multi-stage methods such as MoDe~\cite{zhao2023federated} and FedOSD~\cite{pan2025federated} further require tuning stage-specific learning rates, degradation parameters, and the number of rounds.
FedAU~\cite{gu2024unlearning} introduces additional choices for training duration and for weighting the auxiliary module when merging it with the original model, while FUSED~\cite{zhong2025unlearning} depends on architecture-specific identification of critical layers and associated adapter-tuning hyperparameters.

In contrast, FedQUIT is parameter-light. The distillation temperature is fixed to 1 by default (with theoretical results provided for this configuration, and an extended empirical analysis showing this to be a reasonable choice), the unlearning batch size matches standard local training, and a single local epoch is used by default. The per-sample parameter $v_i=\min_{c} z^g_{i,c}$ in the virtual-teacher construction is data-adaptive, leaving the local unlearning learning rate as the only routine tuning knob.

\section{Background}
In this section, we introduce the essential concepts and notation of FL and FU, followed by a brief overview of KD.
\subsection{Federated Learning and Unlearning}

With $K$ clients holding private datasets $D_k$, FL optimizes
$f(w)=\sum_{k=1}^{K}\tfrac{n_k}{n}F_k(w)$, where $w$ are global parameters, $n_k=|D_k|$,
and $n=\sum_{k=1}^K n_k$. Federated Averaging (FedAvg) proceeds in synchronous rounds:
the server broadcasts $w$ to a selected client subset, each runs $E$ local epochs and
returns updates, and the server updates $w$ via weighted averaging
\cite{mcmahan2017communication,reddi2020adaptive}.

In FL, at round \(t\), a target client \(u\) may request removal of the contribution of a subset \(D_u^{\mathrm{forget}}\subseteq D_u\) (the \emph{forget} data) from the global model \(w_t\).
FU can target \emph{sample}, \emph{client}, \emph{class} \cite{wang2023classdiscriminative}, or \emph{feature} unlearning \cite{gu2025ferrari}.
We focus on client unlearning (\(D_u^{\mathrm{forget}}=D_u\)) and show seamless extension to sample unlearning (\(D_u^{\mathrm{forget}}\subseteq D_u\)).
Let \(D_u^{\mathrm{retain}} = D_u \setminus D_u^{\mathrm{forget}}\); the retain data is \(D_r=\bigcup_k D_k^{\mathrm{retain}}\).
Applying an unlearning algorithm \(\mathcal{U}\) to \(w_t\) produces the post-unlearning global model \(w_t^{\bar{u}} = \mathcal{U}(w_t)\) (\(w^{\bar{u}}\) when \(t\) is clear).
Typically, a \emph{recovery phase} then runs FL rounds on \(D_r\) until model utility is restored at round \(t_{\mathrm{rec}}\), producing the post recovery model \(w_{t_{\mathrm{rec}}}^{\bar{u}}\), as in \cite{halimi2022federated,guo2024not,pan2025federated}.
The target is \(w_{t_{\mathrm{rec}}}^{\bar{u}} \approx w_t^{r}\), where \(w_t^{r}\) is the model retrained from scratch without \(D_u^{\mathrm{forget}}\).
\emph{Efficiency} means minimizing cumulative costs (communication, computation, storage) of running \(\mathcal{U}\) plus recovery; \emph{efficacy} means the post recovery model is approximately indistinguishable from \(w_t^{r}\) \cite{liu2024survey}.

\subsection{Knowledge Distillation}
\label{sec:kd_background}
We consider $C$-class classification. For an input $\mathbf{x}$ with label $y\in\{1,\dots,C\}$,
a network $h(\cdot;w)$ outputs a logit vector (``logits'')
$\mathbf{z}=[z_1,\dots,z_C]\in\mathbb{R}^C$; $z_c$ is its $c$-th component.
The temperature-scaled probabilities are
\[
\mathbf{p}_\tau=\mathrm{softmax}(\mathbf{z}/\tau),\quad
p_{\tau,c}=\frac{\exp(z_c/\tau)}{\sum_{j=1}^C \exp(z_j/\tau)},\quad \tau>0,
\]
with $\mathbf{p}\equiv\mathbf{p}_1$ when $\tau=1$. The one-hot label distribution is
$\mathbf{q}$ with $q_y=1$ and $q_c=0$ for $c\neq y$. The per-sample cross-entropy
(negative log-likelihood) is $\ell(\mathbf{p},y)=-\log p_y$.

\noindent \textbf{KD objective.}
KD transfers knowledge from a fixed \emph{teacher} to a \emph{student} model
by matching softened output distributions \cite{hinton2015distilling, gou2021knowledge}. Let
$\mathbf{z}^S,\mathbf{z}^T\in\mathbb{R}^C$ be student and teacher logits, and define 
$\mathbf{p}^S_\tau=\mathrm{softmax}(\mathbf{z}^S/\tau)$ and $\mathbf{p}^T_\tau=\mathrm{softmax}(\mathbf{z}^T/\tau)$. The canonical KD loss uses the (forward) Kullback–Leibler (KL) divergence:
\[
\mathcal{L}_{\mathrm{KD}}(\mathbf{p}^T_\tau,\mathbf{p}^S_\tau)
\;=\; \tau^2\,\mathrm{KL}\!\big(\mathbf{p}^T_\tau\,\|\,\mathbf{p}^S_\tau\big).
\]

\noindent \textbf{Instantiation in FedQUIT.}
In our setting, the \emph{student} is the client’s local model and the \emph{teacher}
is a \emph{virtual} distribution obtained by modifying the global
model’s logits on forget data (Section~\ref{sec:local_unl}). 
Unless stated otherwise, we set $\tau=1$.

\begin{figure*}[t!]
    \centering
    \includegraphics[width=.98\textwidth]{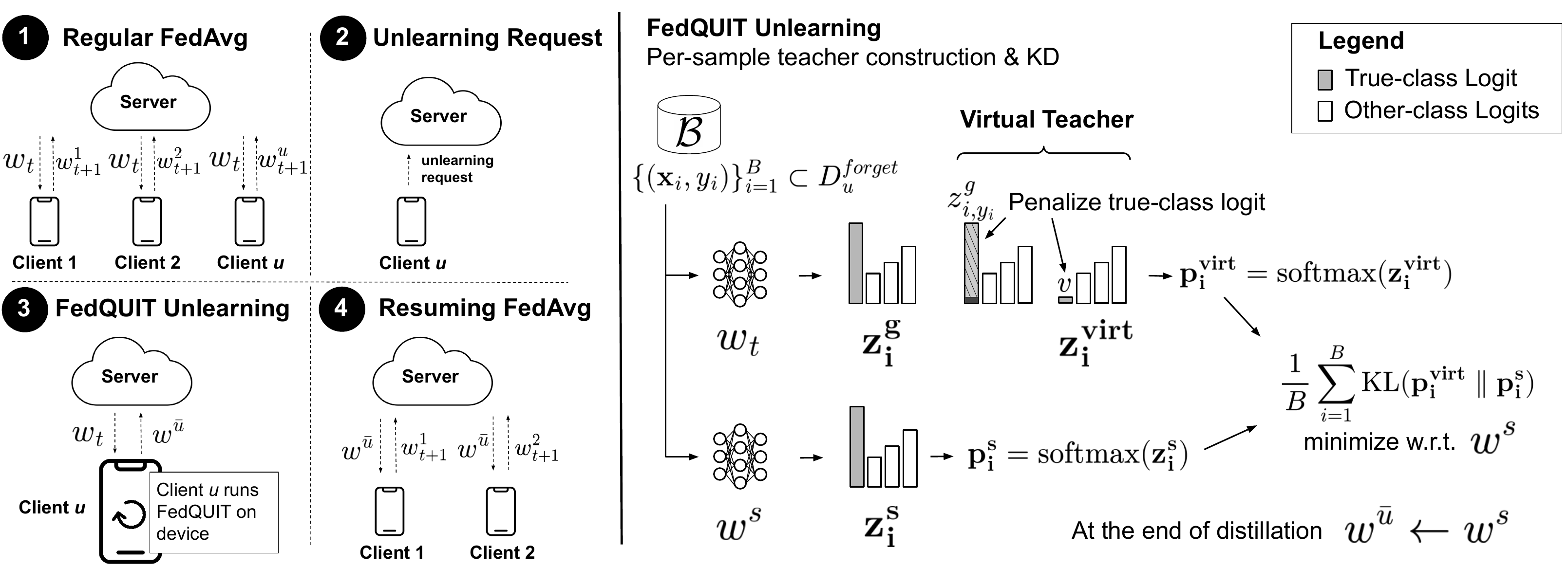}
    \caption{FedQUIT overview. \textbf{Left.} \Circled[fill color=black, inner color=white]{1} Regular training via FedAvg. \Circled[fill color=black, inner color=white]{2} Client $u$ requests unlearning. \Circled[fill color=black, inner color=white]{3} The server initiates a special round. Only client $u$ participates, and runs FedQUIT locally starting from the current global model $w_t$. Client $u$ sends back the unlearned model $w^{\bar{u}}$. \Circled[fill color=black, inner color=white]{4} Regular training resumes from $w^{\bar{u}}$. \textbf{Right.} FedQUIT unlearning phase at client $u$ with forget data $D_u$ (batch size $B$). The global model \( w_t \) computes logits \( \mathbf{z}^g_i \) on a forget sample \( \mathbf{x}_i \). The true-class logit \( \mathbf{z}^{\text{g}}_{y_i} \) is replaced with a fixed or dynamic value \( v \), yielding modified logits \( \mathbf{z}^{\text{virt}}_i \). After applying softmax, the resulting distribution \( \mathbf{p}^{\text{virt}}_i \) serves as a target for the student model \( w_{\bar{u}} \), which is trained to mimic it by minimizing the Kullback-Leibler divergence between \( \mathbf{p}^{\text{virt}}_i \) and the student's output probability \( \mathbf{p}^{\text{s}}_i \).}
    \label{fig:overview}
\end{figure*}

\section{FedQUIT: Unlearning via a Quasi-Competent Virtual Teacher}
\label{sec:method}
In this section, we present FedQUIT, formulated for the client-unlearning setting with a single requesting client, and then describe its natural extension to multiple unlearning requests.

The remainder of this section is organized as follows. 
Section~\ref{sec:protocol} describes the unlearning protocol in a federated setting; Section~\ref{sec:local_unl} details the virtual teacher construction and local unlearning routine; 
Section~\ref{sec:rationale} discusses the underlying intuition and FedAvg compliance; 
Section~\ref{sec:theory} presents the theoretical analysis; Section~\ref{sec:multi_unlearning} extends our method to multiple concurrent unlearning requests.

\subsection{Unlearning Protocol}
\label{sec:protocol}
Our method performs a single-round unlearning phase, after which regular training resumes without the requesting client. During unlearning, the target client’s local model is trained to mimic a \emph{virtual teacher} on its local forget data (explained in detail in Section \ref{sec:local_unl}). 

Figure~\ref{fig:overview} (left) illustrates the unlearning protocol. The client that requested the removal of its contribution performs the unlearning routine itself during a special round, before the regular FL training resumes. During that special round, client $u$ downloads the last version of the global model $w_t$ at the time of the unlearning request, performs FedQUIT locally at the client device, and pushes the unlearned model $w^{\bar{u}}$ back to the server. Then, the server can resume the regular FL training from $w^{\bar{u}}$.

\subsection{Virtual Teacher Construction and KD}
\label{sec:local_unl}

Figure~\ref{fig:overview} (right) and Algorithm~\ref{alg:fedquit} outline the FedQUIT
routine at the target client $u$. The local model acts as the student in a KD
framework; the unlearned model returned to the server will be the snapshot of the student model after the FedQUIT routine. 

At the beginning of the unlearning round $t$, the target client initializes the student with the current global weights, $w^s \leftarrow w_t$ (Alg.~\ref{alg:fedquit}, line~2), as in standard FedAvg rounds. The target client computes the logits of the student and the global model on mini-batches $\{(\mathbf{x}_i,y_i)\}_{i=1}^B \subset D_u^{forget}$ (Alg.~\ref{alg:fedquit}, lines~5–7). Next, the target client constructs a
virtual teacher by modifying the global model’s logits (Alg.~\ref{alg:fedquit},
lines~9–10): letting $\mathbf{z}^g_i=[z^g_{i,1},\dots,z^g_{i,C}]$ be the logits of
$h(\cdot;w_t)$ on $\mathbf{x}_i$, define the virtual-teacher logits
$\mathbf{z}^{\text{virt}}_i=[z^{\text{virt}}_{i,1},\dots,z^{\text{virt}}_{i,C}]$ by
\begin{equation}
\label{eq:virtual_teacher_formula}
z^{\text{virt}}_{i,c}=
\begin{dcases}
v, & c=y_i\\
z^g_{i,c}, & c\neq y_i
\end{dcases}
\quad\text{for } c\in\{1,\dots,C\},
\end{equation}
where $v\in\mathbb{R}$ is a tunable hyperparameter (see \textbf{Choice of $v$} for a parameter-free choice of $v$). The virtual-teacher
probabilities are $\mathbf{p}^{\text{virt}}_i=\mathrm{softmax}(\mathbf{z}^{\text{virt}}_i)$
and the student probabilities are $\mathbf{p}^{s}_i=\mathrm{softmax}(\mathbf{z}^{s}_i)$.
The student mimics the virtual teacher on the forget data by minimizing
\begin{equation}
\label{eq:loss_formula}
\mathcal{L}_{\mathrm{KD}}
= \frac{1}{B}\sum_{i=1}^B \mathrm{KL}\!\big(\mathbf{p}^{\text{virt}}_i \,\|\, \mathbf{p}^{s}_i\big),
\end{equation}
where $\mathrm{KL}$ denotes the Kullback–Leibler divergence (Alg.~\ref{alg:fedquit},
lines~11–13).

\noindent \textbf{Choice of $v$.}
Unless stated otherwise, we set $v_i=\min_{c} z^g_{i,c}$ for each forget sample $x_i$.
This per-sample, data-adaptive rule is parameter-free and 
guarantees $p^{\mathrm{virt}}_{i,y_i}\le 1/C$. The procedure with $v_i=\min_{c} z^g_{i,c}$ is detailed in Appendix B~Algorithm 3. Ablation studies with alternative choices of $v$ are presented in Section~\ref{sec:sensitivity_on_v}.

\subsection{Rationale and Intuition} 
\label{sec:rationale}
When FU is executed locally on the requesting client’s device, the method can access only the target client’s data, while retain data from other clients are unavailable. FedQUIT uses the current global model as an implicit \emph{proxy} for retain knowledge: its soft predictions on the local forget data encode utility-relevant decision structure learned from all clients. Accordingly, we distill this structure by preserving the relative geometry among \emph{non-true} logits (via KD), while explicitly lowering the \emph{true-class} logit to induce forgetting. This design is grounded in two principles: (i) matching non-true logit relationships helps maintain utility \cite{hinton2015distilling}; and (ii) unusually high true-class confidence is a marker of memorization \cite{song2021systematic}. The subsequent analysis formalizes these intuitions by showing that the construction of our \emph{quasi-competent teacher} yields a measurable forgetting signal on the forget set, while—under standard smoothness/boundedness assumptions—keeping the induced update bounded so training can resume without destabilizing FedAvg. Further motivation is provided in Appendix A.

\noindent \textbf{FedAvg compliance.} In standard FedAvg \cite{mcmahan2017communication}, the server sends the global model to active clients at each round; using it for local inference during unlearning adds no extra assumptions. Building a virtual teacher by editing its outputs has negligible cost. FedQUIT relies solely on the local forget data and the current global model (no historical updates, nor public data). 

\subsection{Theoretical Analysis}
\label{sec:theory}

\begin{algorithm*}[t!]
\caption{Local unlearning in \texttt{FedQUIT}.}
\label{alg:fedquit}
\begin{algorithmic}[1]
\Require Global model \(w_t\), forget data \(D_u^{forget}\), unlearning epochs \(E_u\), learning rate \(\eta_u\), batch size \(B\), replacement value for the true-class logit \(v\)
\Ensure Unlearned model parameters \(w^{\bar u}\) to return to the server
\State Client \(u\) receives global model \(w_t\) from the server
\State \textbf{Initialize} student parameters \(w^s \gets w_t\) \Comment{Start from global model}
\For{each local unlearning epoch \(e_u = 1,2,\dots,E_u\)}
  \For{each minibatch \(\mathcal{B}=\{(\mathbf{x}_i,y_i)\}_{i=1}^B \subset D_u^{forget}\)}
    \State \(\mathbf{z}^s \gets h(\mathcal{B}; w^s)\) \Comment{Student logits on batch, \(\mathbf{z}^s \in \mathbb{R}^{B\times C}\)}
    \State \(\mathbf{p}^s \gets \mathrm{softmax}(\mathbf{z}^s)\)
    \State \(\mathbf{z}^g \gets h(\mathcal{B}; w_t)\) \Comment{Global logits on batch, \(\mathbf{z}^g \in \mathbb{R}^{B\times C}\) }
    \State \(\mathbf{z}^{\mathrm{virt}} \gets \mathbf{z}^g\)
    \For{each \(i \in \{1,\dots,B\}\)} 
    \Comment{Modify logits for virtual teacher}
    \State \(\mathbf{z}^{\mathrm{virt}}[i,y_i] \gets v\) \EndFor
    \State \(\mathbf{p}^{\mathrm{virt}} \gets \mathrm{softmax}(\mathbf{z}^{\mathrm{virt}})\)
    \State \(\mathcal{L}_{\mathrm{KD}} \gets \frac{1}{B}\sum_{i=1}^B \mathrm{KL}\!(\mathbf{p}^{\mathrm{virt}}\|\,\mathbf{p}^s\)) \Comment{Temperature \(\tau{=}1\)}
    \State Compute \(\nabla_{w^s}\mathcal{L}_{\mathrm{KD}}\) and update \(w^s \gets \texttt{ClientOpt}(w^s,\nabla_{w^s}\mathcal{L}_{\mathrm{KD}},\eta_u)\)
  \EndFor
\EndFor
\State \(w^{\bar u} \gets w^s\) \Comment{Finalize unlearned snapshot}
\State Client \(u\) sends \(w^{\bar u}\) back to the server
\end{algorithmic}
\end{algorithm*}


We now formalize FedQUIT as a \emph{controlled unlearning perturbation} of the current global model and show how this perturbation simultaneously (a) guarantees a measurable forgetting effect on the forget set and (b) preserves post-unlearning trainability. First, we prove that a KD step toward the \emph{quasi-competent virtual teacher} strictly increases the forget-sample cross-entropy, and we characterize how the strength of this forgetting signal varies monotonically with the teacher parameter $v$ (Lemma~\ref{lem:true-class-penalty}, Proposition~\ref{prop:main-vmin}). Next, we bound the resulting parameter drift $|w^{\bar u}-w_t|$ after $T_u$ local KD steps (Lemma~\ref{lem:main_bounded_perturbation}), which lets us treat FedQUIT as a small warm-start shift. Using this bound, we show that resuming FedAvg on ${k\neq u}$ retains standard convergence behavior, with only an initialization-dependent potential offset controlled by $|w^{\bar u}-w_t|$ (Theorem~\ref{thm:main_fedavg_convergence}). Finally, in Appendix C (Theorem 3) we leverage the same perturbation bound to relate the unlearned model to the ideal retraining solution, making explicit how “closeness to retraining” is governed by the size of the induced perturbation. Assumptions and proofs are deferred to Appendix C.

\begin{lemma}[Penalizing the true-class logit provides a forgetting signal]\label{lem:true-class-penalty}
For a forget sample $(\mathbf{x}_i,y_i)$ let $\mathbf{p}^s_i=\mathrm{softmax}(\mathbf{z}^s_i)$ and
per-sample cross-entropy loss $\ell_i:=-\log p^s_{i,y_i}$.
Define the virtual teacher $\mathbf{p}^{\mathrm{virt}}_i$ as in Eq.~\ref{eq:virtual_teacher_formula} with $v<z^g_{i,y_i}$.
A single KD step
\[
\mathbf{z}^{s+}_i \leftarrow \mathbf{z}^{s}_i - \eta_u\,\nabla_{\mathbf{z}^s_i}\mathrm{KL}\!\big(\mathbf{p}^{\mathrm{virt}}_i\|\mathbf{p}^s_i\big),
\quad \eta_u>0\ \text{small},
\]
satisfies, whenever $p^s_{i,y_i}>p^{\mathrm{virt}}_{i,y_i}$ (as when the student is initialized from the global model, $\mathbf{p}^s_i\approx\mathbf{p}^g_i$),
\[
\Delta\ell_i \;:=\; \ell_i^{+}-\ell_i \;=\; -\log p^{s+}_{i,y_i}+\log p^{s}_{i,y_i} \;>\; 0.
\]
Moving the student toward FedQUIT teacher strictly increases the cross-entropy on the forget data. 
\end{lemma}

\begin{proposition}[Characterization of the forgetting signal]\label{prop:main-vmin}
For a forget sample $(x_i,y_i)$,
let $S_i=\sum_{c\neq y_i} e^{z^g_{i,c}}$ and define
$p^{\mathrm{virt}}_{i,y_i}(v)=\frac{e^{v}}{e^{v}+S_i}$ (non true logits fixed).
Then $p^{\mathrm{virt}}_{i,y_i}(v)$ is strictly increasing in $v$, and the cross-entropy change
$
\Delta\ell_i(v):=\ell_i^{+}-\ell_i
$
(after a small KD step of size $\eta_u>0$, initialized at the global) strictly decreases as $v$ increases.
In particular, choosing $v=\min_{c} z^g_{i,c}$ ensures $p^{\mathrm{virt}}_{i,y_i}\le 1/C$ and yields
$
\Delta\ell_i\big(\min_{c} z^g_{i,c}\big)\ \ge\
\eta_u\,\frac{C}{C-1}\Big(p^g_{i,y_i}-\tfrac{1}{C}\Big)\big(1-p^g_{i,y_i}\big)\;+\;O(\eta_u^2).
$
\end{proposition}

\begin{lemma}[Bounded perturbation induced by FedQUIT]\label{lem:main_bounded_perturbation}
Consider the geometry preserving teacher of Eq.~\ref{eq:virtual_teacher_formula} (true-class logit set to $v$, others fixed) and let the student perform $T_u$ stochastic KD updates with learning rate $\eta_u$ on batches of size $B$, starting from $w_t$. Assume (A4) in Appendix C: there exists $G>0$ such that $\|\nabla_w z(x;w)\|_{\mathrm{op}}\le G$ for all $(x,w)$. Then the returned snapshot satisfies
\[
\|w^{\bar u}-w_t\|
\;\le\; \eta_u\,G\,\sqrt{2}\,T_u .
\]
With default choice $v_i=\min_{c} z^g_{i,c}$, the same bound holds and is typically tighter in practice.
\end{lemma}

\begin{theorem}[FedAvg convergence after resuming from $w^{\bar u}$]
\label{thm:main_fedavg_convergence}
Let
$
F_{\setminus u}(w)
= \sum_{k\neq u}\pi_k\,\mathbb{E}_{(x,y)\sim D_k}[\ell(h(x;w),y)]
$
be the FL objective after removing client $u$, with mixing weights $\pi_k$.
Assume (A5)--(A8) in Appendix C (smoothness, bounded variance, bounded data heterogeneity). Run FedAvg on clients
$\{k\neq u\}$ with any step-size schedule satisfying the standard conditions ensuring FedAvg convergence (e.g.,~\cite{liconvergence}). When initialized at the snapshot $w^{\bar u}$, the FedAvg iterates
converge to a stationary point of $F_{\setminus u}$ with the same rate as
when initialized at $w_t$. The only effect of FedQUIT is the initial
potential shift, with $\|w^{\bar u}-w_t\|$ bounded by
Lemma~\ref{lem:main_bounded_perturbation}:
\begin{equation}
\label{eq:same_convergence}
F_{\setminus u}(w^{\bar u}) - F_{\setminus u}(w_t)
\le
\langle \nabla F_{\setminus u}(w_t),\, w^{\bar u} - w_t \rangle
+\frac{L}{2}\|w^{\bar u} - w_t\|^2 .
\end{equation}
\end{theorem}

\textit{Proof sketch.}
Existing FedAvg convergence results (e.g.,~\cite{liconvergence}) guarantee that the iterates
converge to a stationary point of $F_{\setminus u}$ with a rate
that depends only on the smoothness, variance, heterogeneity constants and the
stepsizes (Assumptions (A5)-(A8) in Appendix C), and not on the particular initialization. Hence, resuming
FedAvg from $w^{\bar u}$ yields the same convergence rate as resuming from
$w_t$; the only difference is the initialization-dependent offset in the
potential $F_{\setminus u}(\cdot)$.

To quantify this offset, use $L$-smoothness of $F_{\setminus u}$ as in Eq. \ref{eq:same_convergence}.
Finally, Lemma~\ref{lem:main_bounded_perturbation} bounds
$\|w^{\bar u}-w_t\|$, implying that FedQUIT induces only a bounded initial
potential shift; plugging this bound into the FedAvg guarantee of, for instance, Li et al. \cite{liconvergence}
completes the proof.
\hfill$\square$

\subsection{Multiple Unlearning Requests}
\label{sec:multi_unlearning}

\begin{algorithm*}[t!]
\caption{Server-side handling of multiple unlearning requests in \texttt{FedQUIT}.}
\label{alg:multi_fedquit}
\begin{algorithmic}[1]
\Require Current global model $w_t$; set of unlearning clients $\mathcal{U}$; local forget dataset $D_u^{\mathrm{forget}}$, $n_u = \sum |D_u^{\mathrm{forget}}|$
\Ensure Unlearned global model $w_t^{\bar {\mathcal{U}}}$

\State \textbf{Server initiates unlearning round}

\For{each $u \in \mathcal{U}$ in parallel}
    \State Send $w_t$ to client $u$
    \State Client $u$ locally executes Algorithm~\ref{alg:fedquit} on $D_u^{\mathrm{forget}}$
    \State Client returns unlearned snapshot $w_t^{\bar u}$
\EndFor

\Statex
\State \textbf{Server performs:}
\Comment FedAvg-like aggregation of unlearned models
\State Compute total weight $N_{\mathcal{U}} \gets \sum_{u\in\mathcal{U}} n_u$
\State
$
w_t^{\bar {\mathcal{U}}} \gets \sum_{u \in \mathcal{U}} \frac{n_u}{N_{\mathcal{U}}} \, w_t^{\bar u}
$

\Statex
\State $w_{t+1} \gets w_t^{\bar {\mathcal{U}}}$
\Comment Initialize next global round
\State Resume standard FedAvg on clients $k \notin \mathcal{U}$

\end{algorithmic}
\end{algorithm*}

\major{Up to now we considered single-client unlearning requests. Now we show that FedQUIT also supports the case where multiple clients request unlearning simultaneously. Algorithm~\ref{alg:multi_fedquit} summarizes the server-side procedure.}

\major{Let $\mathcal{U} \subseteq \{1,\dots,U\}$ denote the set of requesting clients, each wishing to remove the influence of its local forget subset $D_u^{\mathrm{forget}}$. The server can incorporate multiple requests by running a single unlearning round in which all requesting clients participate. Specifically, each $u \in \mathcal{U}$ receives the current global model~$w_t$, executes FedQUIT  locally (Algorithm~\ref{alg:fedquit}, or more specifically, Appendix B Algorithm 3) on its forget data, and returns $w_t^{\bar u}$ to the server.}

\major{After collecting all unlearned snapshots, the server constructs the sanitized global model $w_t^{\bar {\mathcal{U}}}$ by performing a standard FedAvg-like \cite{mcmahan2017communication} weighted aggregation:
\[
w_t^{\bar {\mathcal{U}}} \;=\; \sum_{u \in \mathcal{U}} \frac{n_u}{N_{\mathcal{U}}} \, w_t^{\bar u},
\]
where $n_u = |D_u^{\mathrm{forget}}|$ denotes the size of the forget dataset of client $u$ and $N_{\mathcal{U}} = \sum_{u\in\mathcal{U}} n_u$.  Training then resumes from $w_t^{\bar {\mathcal{U}}}$ using the standard FedAvg procedure while excluding the clients in $\mathcal{U}$ from subsequent participation.}

\major{A potential concern of this parallel formulation is that all requesting clients start from the same global model $w_t$, which still contains the contribution of every client in $\mathcal{U}$. Consequently, each locally unlearned snapshot $w_t^{\bar u}$ is produced while the influence of the other requesting clients is still embedded in the initialization. When these snapshots are later aggregated, one may worry that residual cross-client contributions could be partially reintroduced through the averaging step.}
\major{However, our experimental results show that the penalization and knowledge-distillation mechanism used in the FedQUIT local routine is robust to this aggregation effect. In practice, the local unlearning step sufficiently suppresses the targeted client contribution such that averaging multiple sanitized snapshots does not significantly degrade the unlearning efficacy. The resulting global model consistently achieves performance and forgetting metrics comparable to retraining from scratch, demonstrating that the parallel aggregation preserves retrain-level guarantees.}
\major{As an alternative design, FedQUIT can also support a sequential multi-client unlearning strategy. In this variant, unlearning requests are processed one at a time: after producing a sanitized model for one client, the server updates the global model and then applies FedQUIT again for the next requesting client. While this sequential approach eliminates any residual cross-contribution between requesting clients, it introduces additional computation and communication overhead proportional to the number of requests. In contrast, the parallel procedure described above provides a more efficient solution with negligible loss in unlearning quality, as confirmed by our empirical evaluation.}

\section{Experimental Design}
\label{sec:exp_design}

\begin{table}[t!]
\centering
\adjustbox{max width=\columnwidth}{
    \begin{tabular}{llcccc}
        \toprule
        \textbf{Model} & \textbf{Dataset} & \textbf{Clients} & \textbf{Non-IIDness} & \textbf{Pre-trained} \\
        \midrule
        ResNet-18 & CIFAR-10 & 10 & LDA ($\alpha$=0.3) & No \\ 
        ResNet-18 & CIFAR-100 & 10 & IID & No \\
        ResNet-18 & CIFAR-100 & 10 & LDA ($\alpha$=0.1) & No \\
        MiT-B0 & CIFAR-100 & 10 & LDA ($\alpha$=0.1) & Yes \\
        MiT-B0 & CUB-200 & 10 & LDA ($\alpha$=0.1) & Yes \\
        \major{ResNet-18} & \major{CIFAR-100} & \major{100} & \major{LDA ($\alpha$=0.1)} & \major{No} \\
        \major{MiT-B0} & \major{CUB-200} & \major{100} & \major{LDA ($\alpha$=0.1)}  & \major{Yes} \\
        \major{LSTM}  & \major{Shakespeare} & \major{80} & \major{Contiguous Chunks} & \major{No} \\
        \bottomrule
    \end{tabular}
}
\caption{Federated settings of reported results. Pre-trained column indicates whether the model is trained from scratch or from a pre-trained checkpoint.}
\label{tab:outline_experiments}
\end{table} 

\noindent \textbf{Datasets and data partitioning.} We conduct experiments on federated versions of four datasets, CIFAR-10, CIFAR-100 \cite{Krizhevsky09learningmultiple}, CUB-200 \cite{WelinderEtal2010}, and Tiny-Shakespeare \cite{tiny-sh}. We partition CIFAR-10 among 10 clients, and CIFAR-100 and CUB-200 among both 10 and 100 clients, and consider both IID and non-IID data distributions. We introduce data heterogeneity using distribution-based label skew, controlled via Latent Dirichlet Allocation (LDA) \cite{hsu2019measuring}, with concentration parameters of $\alpha=0.3$ for CIFAR-10 and $\alpha=0.1$ for CIFAR-100 and CUB-200.\footnote{$\alpha\in\{0.1,0.3\}$ provides two client-level label skew regimes, respectively ``moderate'' and ``severe'', to demonstrate that performance trends are not specific to a single data heterogeneity setting.} 
We additionally present results for a next-token prediction task on Tiny-Shakespeare corpus \cite{tiny-sh}, which we partition across 80 clients on contiguous chunks of text.
Table \ref{tab:outline_experiments} reports all the federated settings we consider.

\noindent \textbf{Model and training implementation.} For vision tasks, we employ a standard ResNet-18 \cite{he2016deep} and a vision transformer, MiT-B0 \cite{xie2021segformer}.
Before unlearning, we train ResNet-18 from scratch for 200 FedAvg rounds while fine-tuning MiT-B0 for 50 rounds from a pre-trained checkpoint, with one local epoch per round ($E=1$). The two 100-client configurations (CIFAR-100 with ResNet-18 and CUB-200 with MiT-B0) follow the same protocol but with 600 and 200 rounds respectively. For the Tiny-Shakespeare dataset, we train from scratch for 200 rounds and we use a character-level LSTM architecture with a learned embedding, a 256-unit LSTM encoder, and a classification head. For each setting, we conduct 10 experiments with different target clients and reported mean and standard deviations. 

\noindent \textbf{Baselines.} We use the \textit{Retrain} baseline as gold standard for perfect unlearning, and report values for the Original model (global model before unlearning). We compare our method with six FU SOTA baselines, \Circled[fill color=black, inner color=white]{1} FedEraser \cite{federaser}, \Circled[fill color=black, inner color=white]{2} PGA \cite{halimi2022federated}, \Circled[fill color=black, inner color=white]{3} MoDe \cite{zhao2023federated}, \Circled[fill color=black, inner color=white]{4} FedAU \cite{gu2024unlearning}, \Circled[fill color=black, inner color=white]{5} NoT \cite{khalil2025not}, \Circled[fill color=black, inner color=white]{6} FedOSD \cite{pan2025federated}. We also consider unconstrained gradient ascent as baseline, and we report results in Appendix D. For FedQUIT, we use one epoch for unlearning ($E_u=1$), and $v_i= \min_c z_i^g$. Appendix D reports the description and tuning of the baselines.

\noindent \textbf{Metrics.} We evaluate two perspectives: \emph{efficacy} in forgetting and \emph{efficiency} in recovering model utility. For unlearning efficacy, the most favorable outcome for an approximate unlearning method is to minimize the absolute difference from the gold-standard \textit{Retrain} baseline. We assess:
\begin{itemize}
    \item \textbf{Test Accuracy (Test Acc.)}: Accuracy on the dataset’s held-out test set. 
    \item \textbf{Retain Accuracy (Retain Acc.)}: Accuracy on $D_r$.
    \item \textbf{Forget Accuracy (Forget Acc.)}:
    Accuracy on data $D_u^{forget}$. 
    \item Two \textbf{Membership Inference Attacks (MIAs)}: (1) a confidence-based predictor \cite{song2021systematic} and (2) a loss-based predictor \cite{yeom2018privacy}. See implementation detail in Appendix D, subsection B.
\end{itemize}

For unlearning efficiency, we perform two complementary evaluations:
(a) the resource cost (computation, communication, storage) required to unlearn and recover to retrain-level utility, and
(b) the model performance under fixed efficiency budgets (e.g., fixed communication budgets).
\begin{itemize}
\item \textbf{Efficiency to recover retrain-level utility}: We measure the cumulative \textit{communication} (bytes) and \textit{computation} (FLOPs) consumed during both the unlearning phase and the subsequent recovery phase, following \cite{khalil2025not}. The recovery phase terminates when model utility is restored, i.e., when Test Accuracy matches that of the Retrain baseline. To reduce variance in accuracy trajectories, we consider Test Accuracy averaged over three runs before comparison with Retrain. We additionally record the persistent \textit{storage} (bytes) required across rounds (e.g., historical updates). Further details are provided in Appendix D, subsection C.
\item \textbf{Performance with fixed efficiency budgets}: We also benchmark baseline methods under fixed communication budgets to examine whether forgetting performance degrades once the model has already regained its utility. We express the communication budget as a fraction of the cost of retraining from scratch (defined as 100\%). We evaluate five budget levels: 5\%, 10\%, 20\%, 25\%, and 50\%.
\end{itemize}

\section{Evaluation}
\label{sec:exp_results}
This section discusses the evaluation of FedQUIT. The experimental results shown here aim to answer the following research questions.
\begin{enumerate}
  \item \textbf{Does FedQUIT achieve efficient and effective unlearning (client unlearning, sample unlearning, multiple unlearning requests) under the metrics in Section~\ref{sec:exp_design}?} We compare it against six state-of-the-art baselines, reporting \emph{post-unlearning} and \emph{post-recovery} performance, as well as performance under fixed cost budgets (Section~\ref{sec:fedquit_vs_baselines}). Results indicate that FedQUIT achieves comparable or stronger forgetting efficacy while typically providing substantial improvements in computational and communication efficiency.
  \item \textbf{Is preserving the teacher’s non-true-class rank/geometry essential for distillation?} We ablate the virtual-teacher design across a spectrum from non-informative to fully structure-preserving variants and study the role of distillation temperature (Section~\ref{sec:ablation_teachers}), showing that preserving non-true-class geometry is key to retaining utility and enabling accurate recovery. 
  \item \textbf{How does the FedQUIT-unlearned model differ from the Original and Retrain models?} We analyze predictive-entropy distributions (Section~\ref{sec:predictive_entropy}), showing that FedQUIT closely matches Retrain on forget data while preserving behavior on test data.
  \item \textbf{How sensitive is FedQUIT to key hyperparameters and losses?} We study the tuning of $v$ (Section~\ref{sec:sensitivity_on_v}) and ablate the distillation loss used in Eq.~\ref{eq:loss_formula} (Appendix D, subsection I), finding that the default parameter-free choice is both stable and near-optimal.
\end{enumerate}

\begin{table*}[!t]
\small
\centering
\adjustbox{max width=0.98\textwidth}{
\begin{tabular}{llccccccccc}
\toprule
\multirow{4}{*}{\makecell[l]{\textbf{Setting}}} & \multirow{4}{*}{\makecell[l]{\textbf{Method}}} & \multicolumn{6}{c}{\textbf{Efficacy}} & \multicolumn{3}{c}{\textbf{Efficiency}} \\
\cmidrule(lr){3-8} \cmidrule(lr){9-11}
& & \textbf{Test Acc.} & \textbf{Retain Acc. } & \textbf{Forget Acc. } & \textbf{MIA$_{[Song]}$} & \textbf{MIA$_{[Yeom]}$} & \textbf{Avg.} & \textbf{Communication} & \textbf{Computation} & \textbf{Storage} \\
& & & \textbf{($\Delta$ $\downarrow$)} & \textbf{($\Delta$ $\downarrow$)} & \textbf{($\Delta$ $\downarrow$)} & \textbf{($\Delta$ $\downarrow$)} & \textbf{Gap ($\downarrow$)} & Bytes ($\times$ $\uparrow$) & FLOPs ($\times$ $\uparrow$) & Bytes ($\times$ $\uparrow$) \\
\midrule
\multirow{9}{*}{\makecell[l]{\textbf{CIFAR-100},\\ IID,\\ ResNet-18,\\ $E$=1,\\ 10 Clients}} 
& Original & 59.9$_{\pm 0.0}$ & 79.9$_{\pm 0.1}$ & 79.8$_{\pm 0.7}$ & 78.2$_{\pm 0.6}$ & 71.4$_{\pm 0.6}$ &  & --- & --- & --- \\
& Retrain & 58.3$_{\pm 0.5}$ & 82.7$_{\pm 0.8}$ & 58.0$_{\pm 0.7}$ & 55.6$_{\pm 0.7}$ & 48.8$_{\pm 0.6}$ &  & 1.62e$^{11}$ (1.0×) & 1.35e$^{15}$ (1.0×) & 4.49e$^{07}$ (1.0×) \\
& FedEraser & 58.4$_{\pm 0.5}$ &  \underline{79.5 (3.2$_{\pm 0.7}$)} & \underline{59.5 (1.5$_{\pm 0.7}$)} & \underline{57.6 (2.0$_{\pm 0.5}$)} & \underline{49.9 (1.1$_{\pm 1.1}$)} & 2.0 & 1.62e$^{11}$ (1.0×) & 6.75e$^{14}$ (2.0×) & 8.98e$^{10}$ (0.0005×) \\
& PGA & 59.1$_{\pm 0.6}$ & 76.2 (6.5$_{\pm 0.8}$) & 72.8 (14.8$_{\pm 1.0}$) & 71.8 (16.1$_{\pm 1.8}$) & 63.0 (14.1$_{\pm 1.6}$) & 12.9 & 9.86e$^{09}$ (16.4×) & 8.54e$^{13}$ (15.8×) & 4.94e$^{08}$ (0.0909×) \\
& MoDe & 58.7$_{\pm 0.5}$ & 73.9 (8.8$_{\pm 1.0}$) & 69.0 (11.1$_{\pm 0.7}$) & 70.2 (14.6$_{\pm 2.0}$) & 62.4 (13.5$_{\pm 0.8}$) & 12.0 & 1.75e$^{10}$ (9.3×) & 1.46e$^{14}$ (9.3×) & 8.98e$^{07}$ (0.5×) \\
& FedAU & 59.0$_{\pm 0.5}$ & 75.5 (7.2$_{\pm 0.9}$) & 74.7 (16.8$_{\pm 1.0}$) & 76.3 (20.6$_{\pm 2.1}$) & 61.6 (12.8$_{\pm 2.4}$) & 14.4 & \textbf{2.67e$^{09}$ (60.8×)} & \textbf{2.23e$^{13}$ (60.6×)} & 4.49e$^{07}$ (1.0×) \\
& NoT & 58.9$_{\pm 0.4}$ & 75.2 (7.5$_{\pm 0.6}$) & 70.5 (12.6$_{\pm 0.8}$) & 69.3 (13.7$_{\pm 1.9}$) & 63.3 (14.4$_{\pm 0.8}$) & 12.1 & 5.25e$^{09}$ (30.9×) & 4.39e$^{13}$ (30.8×) & 4.49e$^{07}$ (1.0×) \\
& FedOSD & 58.8$_{\pm 0.5}$ & 76.5 (6.2$_{\pm 0.9}$) & 71.1 (13.1$_{\pm 0.9}$) & 70.9 (15.3$_{\pm 1.9}$) & 61.6 (12.8$_{\pm 1.2}$) & 11.9 & 1.05e$^{10}$ (15.4×) & 8.78e$^{13}$ (15.4×) & 8.98e$^{07}$ (0.5×) \\
& FedQUIT & 58.7$_{\pm 0.4}$ & \textbf{76.6 (6.1$_{\pm 0.8}$)} & \textbf{59.4 (1.4$_{\pm 1.1}$)} & \textbf{60.6 (5.0$_{\pm 1.0}$)} & \textbf{51.8 (3.0$_{\pm 0.8}$)} & \textbf{3.9} & 8.17$e^{09}$ (19.8×) & 6.82$e^{13}$ (19.8×) &  4.49e$^{07}$ (1.0×)\\
\midrule
\multirow{9}{*}{\makecell[l]{\textbf{CIFAR-100},\\ Non-IID,\\ ResNet-18,\\ $E$=1, \\ 10 Clients}} 
& Original & 53.8$_{\pm 0.0}$ & 63.8$_{\pm 0.8}$ & 62.9$_{\pm 6.9}$ & 75.4$_{\pm 7.0}$ & 61.0$_{\pm 7.8}$ &  & --- & --- & --- \\
& Retrain & 51.0$_{\pm 1.4}$ & 64.6$_{\pm 1.2}$ & 33.5$_{\pm 4.5}$ & 44.0$_{\pm 5.5}$ & 32.0$_{\pm 5.2}$ &  & 1.62e$^{11}$ (1.0×) & 1.35e$^{15}$ (1.0×) & 4.49e$^{07}$ (1.0×) \\
& FedEraser & 51.2$_{\pm 0.5}$ & 63.0 (1.6$_{\pm 1.2}$) & \underline{35.7 (2.2$_{\pm 1.6}$)} & \underline{49.1 (3.4$_{\pm 2.0}$)} & \underline{47.4 (3.4$_{\pm 1.5}$)} & 2.7 & 1.62e$^{11}$ (1.0×) & 6.75e$^{14}$ (2.0×) & 8.98e$^{10}$ (0.0005×) \\
& PGA & 51.4$_{\pm 0.6}$ & 62.9 (1.7$_{\pm 1.3}$) & 37.4 (4.3$_{\pm 4.1}$) & 49.2 (6.4$_{\pm 4.9}$) & 36.6 (5.3$_{\pm 5.3}$) & 4.4 & 
5.74e$^{09}$ (28.2×) & 5.10e$^{13}$ (26.5×) & 4.94e$^{08}$ (0.09×) \\
& MoDe & 50.9$_{\pm 1.1}$ & 61.8 (2.8$_{\pm 1.3}$) & 41.9 (8.4$_{\pm 3.8}$) & 55.0 (10.6$_{\pm 4.0}$) & 42.3 (10.3$_{\pm 3.8}$) & 8.0 & 2.19e$^{10}$ (7.4×) & 1.83e$^{14}$ (7.4×) & 8.98e$^{07}$ (0.5×) \\
& FedAU & 51.2$_{\pm 1.2}$ & 60.8 (3.8$_{\pm 1.4}$) & 38.4 (4.5$_{\pm 2.2}$) & 51.7 (6.7$_{\pm 2.5}$) & 38.3 (5.3$_{\pm 1.9}$) & 5.1 & 1.29e$^{10}$ (12.5×) & 1.08e$^{14}$ (12.5×) & 4.49e$^{07}$ (1.0×) \\
& NoT & 51.3$_{\pm 0.2}$ & 57.8 (6.8$_{\pm 1.0}$) & 43.3 (9.9$_{\pm 4.4}$) & 55.7 (11.7$_{\pm 5.7}$) & 44.3 (12.3$_{\pm 5.5}$) & 10.2 & 1.03e$^{10}$ (15.7×) & 8.64e$^{13}$ (15.6×) & 4.49e$^{07}$ (1.0×) \\
& FedOSD & 51.6$_{\pm 0.7}$ & 63.1 (1.5$_{\pm 0.9}$) & 37.3 (3.8$_{\pm 2.0}$) & \textbf{49.0 (5.1$_{\pm 2.4}$)} & 36.0 (4.1$_{\pm 2.1}$) & 3.7 & 9.69e$^{09}$ (16.7×) & 8.10e$^{13}$ (16.7×) & 8.98e$^{07}$ (0.5×) \\
& FedQUIT & 52.0$_{\pm 0.5}$ & \textbf{63.2 (1.4$_{\pm 1.3}$)} &\textbf{34.6 (3.5$_{\pm 2.4}$)} & 49.7 (5.5$_{\pm 2.8}$) & \textbf{33.9 (3.8$_{\pm 2.8}$)} & \textbf{3.6} & \textbf{3.08$e^{09}$ (52.6×)} & \textbf{2.57$e^{13}$ (52.5×)} &  4.49e$^{07}$ (1.0×)\\

\midrule
\multirow{8}{*}{\makecell[l]{\textbf{CIFAR-10},\\ Non-IID,\\ ResNet-18,\\ $E$=1,\\ 10 Clients}} 
& Original & 83.7$_{\pm 0.0}$ & 88.6 $_{\pm 0.3}$ & 88.6$_{\pm 3.9}$ & 84.4$_{\pm 5.4}$ & 81.4$_{\pm 6.3}$ &  & --- & --- & --- \\
& Retrain & 83.5$_{\pm 1.6}$ & 89.0$_{\pm 0.6}$ & 81.1$_{\pm 8.0}$ & 73.1$_{\pm 13.7}$ & 72.5$_{\pm 10.8}$ &  & 1.62e$^{11}$ (1.0×) & 1.35e$^{15}$ (1.0×) & 4.49e$^{07}$ (1.0×) \\
& PGA & 84.0$_{\pm 1.0}$ & 88.7 (0.3$_{\pm 0.6}$) & 84.3 (3.2$_{\pm 4.9}$) & 78.5 (5.4$_{\pm 7.7}$) & 77.0 (4.5$_{\pm 6.0}$) & 3.4 & 9.38e$^{09}$ (17.3×) & 8.14e$^{13}$ (16.6×) & 4.94e$^{08}$ (0.09×) \\
& MoDe & 83.5$_{\pm 1.6}$ & 87.9 (1.1$_{\pm 0.9}$) & 82.3 (2.0$_{\pm 1.7}$) & 72.6 (2.6$_{\pm 1.8}$) & 71.8 (1.5$_{\pm 1.0}$) & 1.8 & 2.40e$^{10}$ (6.7×) & 2.01e$^{14}$ (6.7×) & 8.98e$^{07}$ (0.5×) \\
& FedAU & 83.8$_{\pm 1.4}$ & 89.4 (0.4$_{\pm 0.7}$) & 85.2 (4.0$_{\pm 2.6}$) & 76.9 (4.3$_{\pm 3.7}$) & 73.2 (2.9$_{\pm 2.9}$) & 2.9 & 5.17e$^{09}$ (31.3×) & 4.32e$^{13}$ (31.2×) & 4.49e$^{07}$ (1.0×) \\
& NoT & 83.8$_{\pm 1.2}$ & 83.5 (5.5$_{\pm 0.6}$) & 84.3 (3.1$_{\pm 3.0}$) & 78.0 (4.9$_{\pm 4.5}$) & 75.3 (2.9$_{\pm 3.4}$) & 4.1 & 1.37e$^{10}$ (11.9×) & 1.14e$^{14}$ (11.8×) & 4.49e$^{07}$ (1.0×) \\
& FedOSD & 83.7$_{\pm 1.3}$ & 88.6 (0.4$_{\pm 0.8}$) & 83.6 (2.5$_{\pm 2.2}$) & 76.2 (3.1$_{\pm 2.9}$) & 74.9 (2.4$_{\pm 1.8}$) & 2.1 & 1.05e$^{10}$ (15.4×) & 8.78e$^{13}$ (15.4×) & 8.98e$^{07}$ (0.5×) \\
& FedQUIT & 83.7 $_{\pm 1.2}$ & \textbf{88.8 (0.2$_{\pm 0.7}$)} & \textbf{81.7 (1.4 $_{\pm 1.7}$)} & \textbf{72.1 (1.6 $_{\pm 1.8}$)} & \textbf{71.1 (1.3 $_{\pm 1.5}$)} & \textbf{1.1} & \textbf{3.16$e^{09}$ (51.3×)} & \textbf{2.64$e^{13}$ (51.1×)} &  4.49e$^{07}$ (1.0×)\\

\midrule
\multirow{8}{*}{\makecell[l]{\textbf{CUB-200,}\\ Non-IID,\\ MiT-B0,\\ $E$=1,\\ 10 Clients}} 
& Original & 60.0$_{\pm 0.0}$ & 81.2$_{\pm 1.0}$  & 84.5$_{\pm 10.4}$ & 71.9$_{\pm 12.3}$ & 68.9$_{\pm 10.2}$ &  & -- & -- & --- \\
& Retrain & 56.5$_{\pm 1.0}$ & 82.5$_{\pm 0.7}$ & 34.4$_{\pm 6.2}$ & 22.4$_{\pm 5.5}$ & 19.7$_{\pm 4.8}$ &  & 1.19$e^{10}$ (1.0×) & 3.82$e^{15}$ (1.0×) & 1.33e$^{07}$ (1.0×) \\
& PGA & 57.1$_{\pm 0.4}$ & 82.1 (0.4$_{\pm 0.3}$) & 39.8 (5.4$_{\pm 3.3}$) & 27.8 (5.3$_{\pm 3.2}$) & 26.7 (7.0$_{\pm 3.1}$) & 4.5 & 2.87$e^{09}$ (4.1×) & 9.53$e^{14}$ (4.0×) & 1.46e$^{08}$ (0.0909×) \\
& MoDE & 57.0$_{\pm 1.2}$ & 81.8 (0.7$_{\pm 0.4}$) & 46.6 (12.2$_{\pm 2.8}$) & 29.2 (6.8$_{\pm 2.6}$) & 26.9 (7.2$_{\pm 2.3}$) & 6.7 & 6.67$e^{09}$ (1.8×) & 2.14$e^{15}$ (1.8×) & 2.66e$^{07}$ (0.5×) \\
& FedAU & 56.9$_{\pm 1.2}$ & \textbf{82.3 (0.2$_{\pm 0.5}$)} & \textbf{37.8 (3.4$_{\pm 2.8}$)} & 26.7 (4.3$_{\pm 2.2}$) & 23.0 (3.3$_{\pm 1.6}$) & 2.8 & 6.29$e^{09}$ (1.9×) & 2.01$e^{15}$ (1.9×) & 1.33e$^{07}$ (1.0×) \\
& NoT & 57.0$_{\pm 1.2}$ & 81.2 (1.3$_{\pm 0.6}$) & 54.7 (20.3$_{\pm 5.1}$) & 40.9 (18.6$_{\pm 4.3}$) & 37.5 (17.8$_{\pm 3.4}$) & 14.5 & 2.39$e^{09}$ (5.0×) & 7.65$e^{14}$ (5.0×) & 1.33e$^{07}$ (1.0×) \\
& FedOSD & 57.0$_{\pm 1.0}$ & 80.1 (2.4$_{\pm 0.4}$) & 39.5 (5.1$_{\pm 2.7}$) & 28.0 (5.6$_{\pm 2.5}$) & 24.6 (4.9$_{\pm 2.3}$) & 4.5 & 3.82e$^{09}$ (3.1×) & 1.22e$^{15}$ (3.1×) & 2.66e$^{07}$ (0.5×) \\
& FedQUIT & 57.0$_{\pm 0.9}$ & \textbf{82.3 (0.2$_{\pm 0.1}$)} & 38.0 (3.6$_{\pm 2.7}$) & \textbf{24.6 (2.8$_{\pm 2.5}$)} & \textbf{21.7 (2.0$_{\pm 1.5}$)} & \textbf{2.2} & \textbf{2.20$e^{09}$ (5.4×)} & \textbf{7.05$e^{14}$ (5.4×)} & 1.33e$^{07}$ (1.0×) \\

\midrule
\multirow{7}{*}{\makecell[l]{\textbf{CIFAR-100},\\ Non-IID,\\ ResNet-18,\\ $E$=1, \\ \major{100 Clients}}}
& Original
& \major{35.1$_{\pm 0.6}$}
& \major{36.6$_{\pm 0.7}$}
& \major{49.5$_{\pm 3.8}$}
& \major{59.8$_{\pm 7.0}$}
& \major{48.0$_{\pm 7.5}$}
& & --- & --- & --- \\

& Retrain
& \major{35.1$_{\pm 0.6}$}
& \major{36.8$_{\pm 0.9}$}
& \major{34.2$_{\pm 5.2}$}
& \major{45.2$_{\pm 5.0}$}
& \major{32.5$_{\pm 4.5}$}
& & \major{$5.39$e$^{11}$ (1.0$\times$)} & \major{$4.50$e$^{14}$ (1.0$\times$)} & \major{$4.49$e$^{07}$ (1.0$\times$)} \\

& MoDe
& \major{36.2$_{\pm 0.7}$}
& \major{38.0 (1.2$_{\pm 0.8}$)}
& \major{38.9 (4.7$_{\pm 2.3}$)}
& \major{49.9 (4.7$_{\pm 2.4}$)}
& \major{36.5 (4.0$_{\pm 2.1}$)}
& \major{$3.7$}
& \major{$1.07$e$^{11}$ (5.0$\times$)}
& \major{$8.92$e$^{13}$ (5.0$\times$)}
& \major{$8.98$e$^{07}$ (0.5$\times$)} \\

& FedAU
& \major{36.0$_{\pm 0.8}$}
& \major{34.7 (2.1$_{\pm 0.9}$)}
& \major{39.9 (5.7$_{\pm 2.6}$)}
& \major{50.5 (5.3$_{\pm 2.7}$)}
& \major{38.0 (5.5$_{\pm 2.6}$)}
& \major{$4.7$}
& \major{$3.59$e$^{10}$ (15.0$\times$)}
& \major{$3.00$e$^{13}$ (15.0$\times$)}
& \major{$4.49$e$^{07}$ (1.0$\times$)} \\

& NoT
& \major{36.3$_{\pm 0.6}$}
& \major{34.8 (2.0$_{\pm 0.9}$)}
& \major{40.0 (5.8$_{\pm 2.8}$)}
& \major{51.0 (5.8$_{\pm 2.9}$)}
& \major{38.7 (6.2$_{\pm 3.0}$)}
& \major{$5.0$}
& \major{$4.49$e$^{10}$ (12.0$\times$)}
& \major{$3.75$e$^{13}$ (12.0$\times$)}
& \major{$4.49$e$^{07}$ (1.0$\times$)} \\

& FedOSD
& \major{36.5$_{\pm 0.8}$}
& \major{38.0 (1.2$_{\pm 0.8}$)}
& \major{39.9 (5.7$_{\pm 2.5}$)}
& \major{50.2 (5.0$_{\pm 2.5}$)}
& \major{37.8 (5.3$_{\pm 2.4}$)}
& \major{$4.3$}
& \major{$1.26$e$^{10}$ (42.9$\times$)}
& \major{$1.05$e$^{13}$ (42.9$\times$)}
& \major{$8.98$e$^{07}$ (0.5$\times$)} \\

& FedQUIT
& \major{36.5$_{\pm 0.7}$}
& \major{\textbf{37.9 (1.1$_{\pm 0.7}$)}}
& \major{\textbf{36.1 (1.9$_{\pm 1.5}$)}}
& \major{\textbf{46.8 (1.6$_{\pm 1.4}$)}}
& \major{\textbf{34.8 (2.3$_{\pm 1.6}$)}}
& \major{\textbf{1.7}}
& \major{\textbf{$7.27$e$^{09}$ (74.1$\times$)}}
& \major{\textbf{$6.08$e$^{12}$ (74.1$\times$)}}
& \major{$4.49$e$^{07}$ (1.0$\times$)} \\ 

\midrule

\multirow{7}{*}{\makecell[l]{\textbf{CUB-200},\\ Non-IID,\\ MiT-B0,\\ $E$=1, \\ \major{100 Clients}}}
& Original
& \major{45.3$_{\pm 0.9}$}
& \major{57.4$_{\pm 0.8}$}
& \major{72.0$_{\pm 5.5}$}
& \major{60.0$_{\pm 8.0}$}
& \major{57.0$_{\pm 7.5}$}
& & --- & --- & --- \\

& Retrain
& \major{43.8$_{\pm 1.0}$}
& \major{56.7$_{\pm 0.9}$}
& \major{41.3$_{\pm 6.3}$}
& \major{29.0$_{\pm 5.5}$}
& \major{26.0$_{\pm 4.8}$}
& & \major{$5.31$e$^{10}$ (1.0$\times$)} & \major{$2.04$e$^{14}$ (1.0$\times$)} & \major{$1.33$e$^{07}$ (1.0$\times$)} \\

& MoDe
& \major{44.4$_{\pm 1.1}$}
& \major{57.0 (0.3$_{\pm 0.5}$)}
& \major{43.6 (2.3$_{\pm 1.8}$)}
& \major{31.6 (2.6$_{\pm 2.0}$)}
& \major{29.6 (3.6$_{\pm 2.3}$)}
& \major{$2.2$}
& \major{$3.16$e$^{10}$ (1.7$\times$)}
& \major{$1.21$e$^{14}$ (1.7$\times$)}
& \major{$2.66$e$^{07}$ (0.5$\times$)} \\

& FedAU
& \major{44.2$_{\pm 1.0}$}
& \major{56.8 (0.1$_{\pm 0.4}$)}
& \major{43.6 (2.3$_{\pm 1.8}$)}
& \major{31.6 (2.6$_{\pm 2.0}$)}
& \major{29.6 (3.6$_{\pm 2.3}$)}
& \major{$2.1$}
& \major{$7.97$e$^{09}$ (6.7$\times$)}
& \major{$3.06$e$^{13}$ (6.7$\times$)}
& \major{$1.33$e$^{07}$ (1.0$\times$)} \\

& NoT
& \major{44.5$_{\pm 1.2}$}
& \major{56.5 (0.2$_{\pm 0.5}$)}
& \major{64.0 (22.7$_{\pm 4.5}$)}
& \major{52.0 (23.0$_{\pm 4.8}$)}
& \major{50.0 (24.0$_{\pm 5.0}$)}
& \major{$17.5$}
& \major{$4.25$e$^{09}$ (12.5$\times$)}
& \major{$1.63$e$^{13}$ (12.5$\times$)}
& \major{$1.33$e$^{07}$ (1.0$\times$)} \\

& FedOSD
& \major{44.5$_{\pm 1.0}$}
& \major{\textbf{56.7 (0.0$_{\pm 0.4}$)}}
& \major{43.0 (1.7$_{\pm 1.5}$)}
& \major{31.0 (2.0$_{\pm 1.8}$)}
& \major{29.0 (3.0$_{\pm 2.0}$)}
& \major{$1.7$}
& \major{$4.25$e$^{09}$ (12.5$\times$)}
& \major{$1.63$e$^{13}$ (12.5$\times$)}
& \major{$2.66$e$^{07}$ (0.5$\times$)} \\

& FedQUIT
& \major{44.5$_{\pm 1.1}$}
& \major{56.9 (0.2$_{\pm 0.4}$)}
& \major{\textbf{39.7 (1.6$_{\pm 1.4}$)}}
& \major{\textbf{27.7 (1.3$_{\pm 1.3}$)}}
& \major{\textbf{25.7 (0.3$_{\pm 1.0}$)}}
& \major{\textbf{0.9}}
& \major{\textbf{$8.23$e$^{08}$ (64.5$\times$)}}
& \major{\textbf{$3.16$e$^{12}$ (64.5$\times$)}}
& \major{$1.33$e$^{07}$ (1.0$\times$)} \\

\bottomrule
\end{tabular}
}
\caption{Post-recovery performance of FedQUIT vs. baselines. Metrics are \textit{mean ($\Delta$ $\pm$ std)}, where $\Delta$ is the average absolute gap with \textit{Retrain}. Lower $\Delta$ corresponds to better unlearning. The \textit{Avg. Gap} column reports the average across $\Delta$. 
For efficiency, higher $\times$ means greater reduction vs. \textit{Retrain}.
When FedEraser \cite{federaser} ranks first, it is underlined to note efficiency limits.}
\label{table:full_results}
\end{table*}

\subsection{Evaluation of Unlearning Efficacy and Efficiency}
\label{sec:fedquit_vs_baselines}
\noindent \textbf{Post-recovery Results.} Table~\ref{table:full_results} reports post-recovery performance (i.e., once each method’s test accuracy matches \emph{Retrain}). \textbf{Findings.} Under IID settings (first row), FedQUIT is the only approach that substantially forgets the target client without storing a persistent history of client updates (as in FedEraser), while reducing cumulative communication and computation by $\sim\!20\times$ relative to \emph{Retrain}. Under non-IID settings with both 10 and 100 clients, FedQUIT closely matches \emph{Retrain}, attaining the lowest or second-lowest Avg.\ Gap on efficacy metrics and the largest or second-largest efficiency gains, without the storage burden required by FedEraser. We validate our results also using more local epochs ($E=10$), results included in Appendix Tables X and XI.

\noindent \textbf{Fixed-budget Evaluation.} 
To examine behavior beyond recovery, Table~\ref{table:fixed_budget_50} 
compares all methods under a large fixed communication budget set to 50\% of the cost of retraining from scratch (additional budgets are in Appendix D, subsection E and Appendix Table XIII). This controls for the possibility that forgetting degrades after utility is restored. \textbf{Findings.} FedQUIT consistently achieves the smallest Forget Acc.\ gaps and often higher Test and Retain Acc.\ than baselines at the same budget, indicating it recovers utility while preserving selective forgetting; the same trend holds across all considered budgets.

\begin{table*}[t]
\centering
\small
\adjustbox{max width=\textwidth}{
\begin{tabular}{l l c c c c c c}
\toprule
 & & \multicolumn{3}{c}{\major{\textbf{IID}}} 
   & \multicolumn{3}{c}{\major{\textbf{Non-IID}}} \\
\cmidrule(lr){3-5} \cmidrule(lr){6-8}
\major{\textbf{Budget}} 
  & \major{\textbf{Method}} 
  & \major{\textbf{Test Acc. ($\Delta$)}} 
  & \major{\textbf{Retain Acc. ($\Delta$)}} 
  & \major{\textbf{Forget Acc. ($\Delta$)}} 
  & \major{\textbf{Test Acc. ($\Delta$)}} 
  & \major{\textbf{Retain Acc. ($\Delta$)}} 
  & \major{\textbf{Forget Acc. ($\Delta$)}} \\
\midrule

\major{100\%} 
 & Retrain 
 & 58.31 & 82.72 & 58.02 
 & 51.02 & 64.61 & 33.52 \\
\midrule

\multirow{7}{*}{\major{50\%}}
 & FedEraser 
   & 45.48 (\textcolor{red}{-12.83}) 
   & 56.28 (\textcolor{red}{-26.44}) 
   & 45.42 (\textcolor{red}{-12.60})
   & 37.76 (\textcolor{red}{-13.26}) 
   & 41.32 (\textcolor{red}{-23.29}) 
   & 19.99 (\textcolor{red}{-13.53}) \\

 & PGA 
   & 61.09 (+2.78) 
   & 89.95 (+7.23) 
   & 68.83 (+10.81)
   & 55.20 (+4.18) 
   & 73.31 (+8.70) 
   & 40.49 (+6.97) \\

 & MoDe       
   & 61.44 (+3.13)
   & 91.98 (+9.26)
   & 68.50 (+10.48)
   & 54.95 (+3.93)
   & 72.45 (+7.84)
   & 41.18 (+7.66) \\

 & FedAU      
   & 61.57 (+3.26)
   & 91.03 (+8.31)
   & 66.66 (+8.64)
   & 55.38 (+4.36)
   & 74.42 (+9.81)
   & 41.88 (+8.36) \\

 & NoT        
   & 61.05 (+2.74)
   & 90.59 (+7.87)
   & 66.36 (+8.34)
   & 55.10 (+4.08)
   & 73.27 (+8.66)
   & 40.39 (+6.87) \\

 & FedOSD 
   & 61.25 (+2.94) 
   & 90.32 (+7.60) 
   & 66.88 (+8.86)
   & 55.30 (+4.28) 
   & 74.14 (+9.53) 
   & 39.76 (+6.24) \\

 & FedQUIT    
   & \textbf{61.67 (+3.36)}
   & \textbf{91.52 (+8.80)}
   & \textbf{62.58 (+4.56)}
   & \textbf{55.48 (+4.46)}
   & \textbf{74.47 (+9.86)}
   & \textbf{37.51 (+3.99)} \\

\bottomrule
\end{tabular}
}
\caption{\major{Comparison at fixed communication budget of 50\%
($8.10e^{10}$ bytes). 
Values in parentheses indicate the gap (negative gap in red) with \textit{Retrain}
for the corresponding setting (IID or Non-IID). Setting: CIFAR-100, ResNet-18, 10 clients.}}
\label{table:fixed_budget_50}
\end{table*}

 \noindent \textbf{Post-unlearning Results.} 
 We quantify the utility model's drop caused by the unlearning phase alone (before recovery). Table~\ref{tab:mini_table_after_unlearning} reports Test Accuracy immediately after unlearning and the unlearning-only cost; full results are in Appendix Table XI. \textbf{Findings.} Under IID, FedQUIT achieves the second-highest post-unlearning Test Accuracy (after MoDe), while avoiding MoDe’s multi-round unlearning cost, which is typically $\sim\!150\times$ higher FLOPs/bytes in our settings (Appendix D, subsection A). Other baselines are less selective on IID and often collapse utility. Under non-IID, FedOSD (and for transformers, MoDe) preserves utility slightly better but at much higher cost—typically $\sim\!90\times$ more FLOPs/bytes—whereas FedQUIT remains close in Test Accuracy at orders-of-magnitude lower cost. NoT has near-zero unlearning cost but markedly degrades utility (shifting cost to recovery), and PGA/FedAU similarly harm utility during unlearning.

\begin{table*}[t!]
\centering
\adjustbox{max width=\textwidth}{
\begin{tabular}{lccccccccccccc}
\toprule
& \multicolumn{3}{c}{\textbf{CIFAR-100, IID, ResNet-18}} 
& \multicolumn{3}{c}{\textbf{CIFAR-100, N-IID, ResNet-18}} 
& \multicolumn{3}{c}{\textbf{CIFAR-100, N-IID, MiT-B0}} 
& \multicolumn{3}{c}{\textbf{CUB-200, N-IID, MiT-B0}} \\
\cmidrule(lr){2-4}\cmidrule(lr){5-7}\cmidrule(lr){8-10}\cmidrule(lr){11-13}
\textbf{Method} 
& \textbf{Test Acc.} & \textbf{Bytes} & \textbf{FLOPs}
& \textbf{Test Acc.} & \textbf{Bytes} & \textbf{FLOPs}
& \textbf{Test Acc.} & \textbf{Bytes} & \textbf{FLOPs}
& \textbf{Test Acc.} & \textbf{Bytes} & \textbf{FLOPs} \\
\midrule
Retrain     
& $58.3_{\pm 0.5}$ & 1.62$e^{11}$ & 1.62$e^{15}$           
& $51.0_{\pm 1.4}$ & 1.62$e^{11}$ & 1.62$e^{15}$    
& $73.3_{\pm 0.8}$ & 1.19$e^{10}$ & 3.82$e^{15}$    
& $56.5_{\pm 1.0}$ & 1.19$e^{10}$ & 3.82$e^{15}$ \\
PGA         
& $1.0_{\pm 0.2}$  & 8.98e$^{07}$ & 3.75e$^{12}$ 
& $1.0_{\pm 0.2}$  & 8.98e$^{07}$ & 3.75e$^{12}$
& $10.1_{\pm 0.2}$ & 2.66e$^{07}$ & 4.25e$^{13}$
& $5.4_{\pm 0.2}$  & 2.66e$^{07}$ & 4.25e$^{13}$ \\
MoDe$^\dag$ 
& $36.8_{\pm 5.6}$ & 1.38e$^{10}$ & 1.16e$^{14}$ 
& $17.5_{\pm 3.6}$ & 1.38e$^{10}$ & 1.16e$^{14}$
& $66.6_{\pm 2.1}$ & 4.09e$^{09}$ & 1.31e$^{15}$
& $43.4_{\pm 2.2}$ & 4.09e$^{09}$ & 1.31e$^{15}$ \\
FedAU       
& $4.2_{\pm 1.3}$  & 8.21e$^{07}$ & 1.84e$^{12}$ 
& $4.8_{\pm 1.5}$  & 8.21e$^{07}$ & 1.84e$^{12}$
& $1.8_{\pm 1.1}$  & 1.03e$^{07}$ & 4.61e$^{10}$
& $0.8_{\pm 0.2}$  & 1.03e$^{07}$ & 4.61e$^{10}$ \\
NoT$^*$         
& $12.2_{\pm 0.0}$ & 0         & 0            
& $9.5_{\pm 0.0}$  & 0          & 0
& $41.3_{\pm 0.0}$ & 0          & 0
& $25.3_{\pm 0.0}$ & 0          & 0 \\
FedOSD$^\dag$
& $4.5_{\pm 1.6}$  & 8.08e$^{09}$ & 6.75e$^{13}$ 
& $49.5_{\pm 0.9}$ & 8.08e$^{09}$ & 6.75e$^{13}$
& $58.8_{\pm 2.1}$ & 2.39e$^{09}$ & 7.65e$^{14}$
& $52.7_{\pm 1.2}$ & 2.39e$^{09}$ & 7.65e$^{14}$ \\
FedQUIT     
& $25.5_{\pm 1.1}$ & 8.98e$^{07}$ & 7.50e$^{11}$ 
& $45.2_{\pm 1.0}$ & 8.98e$^{07}$ & 7.50e$^{11}$
& $50.4_{\pm 1.6}$ & 2.66e$^{07}$ & 8.50e$^{12}$
& $47.8_{\pm 1.9}$ & 2.66e$^{07}$ & 8.50e$^{12}$ \\
\bottomrule
\end{tabular}
}
\caption{Post-unlearning Test Acc. and cost (Bytes, FLOPs). $^\dag$: multi-round unlearning with active target client; $^*$: identical unlearned model for any target client (std = 0). 
}
\label{tab:mini_table_after_unlearning}
\end{table*}

\noindent \textbf{Sample-Unlearning Results.}
We evaluate three forget-set sizes: 50\% and 10\% of local data drawn at random, and, as an extreme case, 1\% drawn from the least-represented local classes. Table \ref{table:sample_unlearning_main} reports the results for the 1\% setting, while Appendix D, subsection G, provides complete results for all configurations, under both Non-IID and IID scenarios. \textbf{Findings.} In this extreme setting (1\% forget data), FedQUIT outperforms all baselines, showing both lower degradation after unlearning and more accurate forgetting after recovery, while also requiring substantially lower computation and communication costs ($\approx 4\times$). Similar trends hold for the less extreme cases (see Appendix D, subsection G).
\begin{table*}[t!]

\centering
\adjustbox{max width=\textwidth}{
\begin{tabular}{lcccccccc}
\toprule
\major{\textbf{Method}} &
\multicolumn{2}{c}{\major{\textbf{Test Accuracy}}} &
\multicolumn{2}{c}{\major{\textbf{Retain Accuracy}}} &
\multicolumn{2}{c}{\major{\textbf{Forget Accuracy}}} &
\major{\textbf{Comm.}} &
\major{\textbf{Comp.}} \\
\cmidrule(lr){2-3} \cmidrule(lr){4-5} \cmidrule(lr){6-7}

& \major{\textbf{After Unlearning}} &
  \major{\textbf{After Recovery}} &
  \major{\textbf{After Unlearning}} &
  \major{\textbf{After Recovery}} &
  \major{\textbf{After Unlearning}} &
  \major{\textbf{After Recovery}} &
  \major{(Bytes $\downarrow$)} &
  \major{(FLOPs $\downarrow$)} \\
\midrule

Original &
\major{51.81} & \major{51.81} &
\major{63.79} & \major{63.79}&
\major{66.67} & \major{66.67}&
\major{} & \major{} \\

Retrain &
\major{50.84 $\pm$ 0.3} & \major{50.84 $\pm$ 0.3}&
\major{64.82 $\pm$ 0.3} & \major{64.82 $\pm$ 0.3}&
\major{55.00 $\pm$ 1.5} & \major{55.00 $\pm$ 1.5}&
\major{$1.80 e^{11}$} & \major{$1.35 e^{15}$} \\

MoDe &
\major{42.20 (8.64 $\pm$ 0.3)} & \major{51.53 (0.69 $\pm$ 0.3)} &
\major{36.30 (28.52 $\pm$ 0.4)} & \major{63.20 (1.62 $\pm$ 0.3)} &
\major{8.51 (46.49 $\pm$ 1.9)} & \major{58.70 (3.70 $\pm$ 1.7)} &
\major{$6.19 e^{9}$} & \major{$5.18 e^{13}$} \\

FedAU &
\major{24.50 (26.34 $\pm$ 0.4)} & \major{51.10 (0.26 $\pm$ 0.3)} &
\major{23.50 (41.32 $\pm$ 0.4)} & \major{63.60 (1.22 $\pm$ 0.3)} &
\major{0.00 (55.00 $\pm$ 2.2)} & \major{59.00 (4.00 $\pm$ 1.8)} &
\major{$5.39 e^{9}$} & \major{$4.05 e^{13}$} \\

NoT &
\major{9.70 (41.14 $\pm$ 0.4)} & \major{51.50 (0.66 $\pm$ 0.3)} &
\major{9.10 (55.72 $\pm$ 0.4)} & \major{64.40 (0.42 $\pm$ 0.3)} &
\major{15.12 (39.88 $\pm$ 2.3)} & \major{58.67 (3.67 $\pm$ 1.9)} &
\major{$7.18 e^{9}$} & \major{$5.41 e^{13}$} \\

FedOSD &
\major{49.30 (1.54 $\pm$ 0.2)} & \major{\textbf{51.93 (1.09 $\pm$ 0.2)}} &
\major{60.51 (4.31 $\pm$ 0.3)} & \major{63.95 (0.87 $\pm$ 0.3)} &
\major{\textbf{60.00 (5.00 $\pm$ 1.8)}} & \major{57.40 (2.40 $\pm$ 1.5)} &
\major{$4.04 e^{9}$} & \major{$3.38 e^{13}$} \\

FedQUIT &
\major{\textbf{50.79 (0.05 $\pm$ 0.3)}} & \major{52.00 (1.16 $\pm$ 0.2)} &
\major{\textbf{61.20 (3.62 $\pm$ 0.3)}} &
\major{\textbf{64.53 (0.29 $\pm$ 0.3)}} &
\major{41.13 (13.87 $\pm$ 1.3)} &
\major{\textbf{56.67 (1.67 $\pm$ 1.0)}} &
\major{\textbf{9.87 $e^{8}$}} &
\major{\textbf{7.51 $e^{12}$}} \\

\bottomrule
\end{tabular}}
\caption{\major{Post-unlearning and post-recovery performance for \textbf{sample unlearning} on CIFAR-100 (Non-IID, 1\% forget data, ResNet-18, 10 clients). Gaps vs.\ \textit{Retrain} in parenthesis with added $\pm$ std.}}
\label{table:sample_unlearning_main}
\end{table*}

\noindent \textbf{Multiple Unlearning Results.}
\major{Table~\ref{table:multiple_unlearning_after_rec} and Appendix Table XVIII report, respectively, the after-recovery and after-unlearning results when two clients request unlearning simultaneously, across the two settings indicated on the left column. We do not include FedOSD \cite{pan2025federated} because it does not naturally support multiple unlearning requests.
\textbf{Findings.} FedQUIT achieves the best retained performance immediately after unlearning (higher Test and Retain accuracy), provides the most accurate post-recovery unlearning metrics, and outperforms all baselines in terms of efficiency (lower communication and computation costs to recover utility).}
\begin{table*}[!t]
\centering
\adjustbox{max width=\textwidth}{
\begin{tabular}{llcccccc}
\toprule
\multirow{4}{*}{\makecell[l]{\major{\textbf{Setting}}}} 
& \multirow{4}{*}{\makecell[l]{\major{\textbf{Method}}}} 
& \multicolumn{4}{c}{\major{\textbf{Efficacy}}} 
& \multicolumn{2}{c}{\major{\textbf{Efficiency}}} \\
\cmidrule(lr){3-6} \cmidrule(lr){7-8}
& 
& \major{\textbf{Test Acc.}} 
& \major{\textbf{Forget Acc. ($\Delta$ $\downarrow$)}} 
& \major{\textbf{MIA$_{[Song]}$ ($\Delta$ $\downarrow$)}} 
& \major{\textbf{MIA$_{[Yeom]}$ ($\Delta$ $\downarrow$)}} 
& \major{\textbf{Communication}} 
& \major{\textbf{Computation}} \\
& 
& 
& 
& 
& 
& \major{Bytes ($\times$ $\uparrow$)} 
& \major{FLOPs ($\times$ $\uparrow$)} \\
\midrule

\multirow{6}{*}{\makecell[l]{\textbf{\major{Multiple Unlearning}},\\ CIFAR-100, Non-IID,\\ ResNet-18, $E$=1, \\ 10 Clients}}
& Original     & 53.8$_{\pm 0.0}$ & 63.1$_{\pm 6.8}$ & 76.0$_{\pm 7.6}$ & 61.1$_{\pm 8.3}$ & --- & --- \\
& Retrain      & 47.3$_{\pm 2.1}$ & 30.1$_{\pm 3.3}$ & 39.7$_{\pm 3.9}$ & 27.9$_{\pm 3.1}$ & 1.44e$^{11}$ & 1.20e$^{15}$\\
& MoDe         & 48.2$_{\pm 2.0}$ & 35.7 (5.6$_{\pm 2.0}$) & 44.7 (5.0$_{\pm 2.7}$) & 32.2 (4.1$_{\pm 3.3}$) & 1.74e$^{10}$ (8.3×) & 1.36e$^{14}$ (8.8×)\\
& FedAU        & 48.3$_{\pm 2.3}$ & 35.6 (5.7$_{\pm 2.4}$) & 45.6 (6.4$_{\pm 3.6}$) & 33.6 (5.7$_{\pm 4.1}$) & 7.90e$^{09}$ (18.2×) & 6.60e$^{13}$ (18.2×) \\
& NoT          & 48.1$_{\pm 2.2}$ & 39.2 (9.1$_{\pm 2.8}$) & 50.5 (11.3$_{\pm 3.4}$) & 41.8 (13.9$_{\pm 4.5}$) & 4.74e$^{09}$ (30.4×) & 3.96e$^{13}$ (30.3×) \\
& FedQUIT      & 48.5$_{\pm 1.9}$ 
               & \textbf{32.7 (2.6$_{\pm 1.2}$)} 
               & \textbf{41.9 (2.2$_{\pm 1.1}$)} 
               & \textbf{30.1 (2.2$_{\pm 2.0}$)} 
               & \textbf{2.33e$^{09}$ (61.7×)} 
               & \textbf{1.95e$^{13}$ (61.5×)} \\

\midrule

\multirow{6}{*}{\makecell[l]{\textbf{\major{Multiple Unlearning}},\\ CIFAR-100, Non-IID,\\ MiT-B0, $E$=1, \\ 10 Clients}}
& Original     & 75.0$_{\pm 0.0}$ & 84.6$_{\pm 5.3}$ & 76.8$_{\pm 9.0}$ & 73.8$_{\pm 1.4}$ & --- & --- \\
& Retrain      & 70.9$_{\pm 1.3}$ & 54.3$_{\pm 3.5}$ & 45.4$_{\pm 3.6}$ & 43.5$_{\pm 1.6}$ & 1.06e$^{10}$  & 3.40e$^{15}$  \\
& MoDe         & 71.5$_{\pm 0.6}$ 
               & 59.4 (5.1$_{\pm 1.3}$) 
               & 50.4 (5.0$_{\pm 1.8}$)
               & 47.6 (4.1$_{\pm 2.9}$) 
               & 5.68e$^{09}$ (1.8×) 
               & 1.69e$^{15}$ (2.0×) \\

& FedAU        & 71.4$_{\pm 0.9}$ & 58.9 (4.6$_{\pm 2.3}$) & 50.1 (4.7$_{\pm 2.3}$) & 47.2 (3.7$_{\pm 1.9}$) & 3.19e$^{09}$ (3.3×) & 1.02e$^{15}$ (3.3×) \\
& NoT          & 71.4$_{\pm 0.9}$ & 69.9 (15.6$_{\pm 6.3}$) & 62.5 (16.6$_{\pm 8.9}$) & 59.0 (15.5$_{\pm 7.5}$) & 6.80e$^{08}$ (15.6×) & 2.18e$^{14}$ (15.6×) \\
& FedQUIT      & 71.6$_{\pm 0.7}$ 
               & \textbf{56.3 (2.0$_{\pm 1.0}$)} 
               & \textbf{47.1 (1.7$_{\pm 0.9}$)}
               & \textbf{45.3 (1.8$_{\pm 1.0}$)} 
               & \textbf{4.78e$^{08}$ (22.2×)} 
               & \textbf{1.53e$^{14}$ (22.2×)} \\

\bottomrule
\end{tabular}
}
\caption{\major{Post-recovery performance for \textbf{multiple unlearning requests} (two clients out of ten). The results are averaged over five different pairs of clients working as requesting clients. Results are expressed as \textit{mean metric value (mean $\Delta$ $\pm$ standard deviation)}, with $\Delta$ representing the average absolute difference with \textit{Retrain}. Lower $\Delta$ corresponds to better unlearning.}}
\label{table:multiple_unlearning_after_rec}
\end{table*}
\captionsetup{labelfont={color=black}}

\noindent \textbf{Next-token Prediction Results.}
Table~\ref{table:lstm} reports the results for the Tiny-Shakespeare setting, framed as a character-level next-token prediction task.
\textbf{Findings.}
Consistently with the trends observed in the other configurations 
(e.g., Table ~\ref{table:full_results}, Appendix Table XI, and Appendix 
Table XII), FedQUIT emerges as the most 
efficient baseline for recovering model utility, requiring the lowest communication 
and computation budgets to return to retrain-level test accuracy. 
This efficiency stems from the characteristics of the FedQUIT model immediately 
after unlearning (left part of Table~\ref{table:lstm}): it exhibits selective 
forgetting that leads to only minor degradation in test and retain accuracy, 
while inducing a larger drop in forget accuracy. This selective degradation makes 
the subsequent recovery procedure considerably lighter. Moreover, FedQUIT achieves the most accurate forgetting among the compared baselines, 
showing the smallest absolute gap from the \textit{Retrain} forget accuracy, 
confirming its strong ability to remove target information while quickly recovering overall utility.

\begin{table*}[t]
\centering
\adjustbox{max width=\textwidth}{
\begin{tabular}{ l c c c c c c c c}
\toprule
  & \multicolumn{3}{c}{\textbf{After Unlearning}} 
 & \multicolumn{3}{c}{\textbf{After Recovery}}
 & \multicolumn{2}{c}{\textbf{Efficiency}} \\
\cmidrule(lr){2-4}\cmidrule(lr){5-7}\cmidrule(lr){8-9}
 \textbf{Method} 
 & \textbf{Test Acc.} 
 & \textbf{Retain Acc.} 
 & \textbf{Forget Acc.}
 & \textbf{Test Acc.} 
 & \textbf{Retain Acc. ($\Delta$ $\downarrow$)} 
 & \textbf{Forget Acc. ($\Delta$ $\downarrow$)}
 & \textbf{Comm. (Bytes)} & \textbf{Comp. (FLOPs)} \\
\midrule


Original 
 & 57.93$_{\pm 0.2}$ 
 & 61.56$_{\pm 0.3}$ 
 & 66.42$_{\pm 0.2}$
 & 57.93$_{\pm 0.2}$ 
 & 61.56 $_{\pm 0.3}$ 
 & 66.42 $_{\pm 0.3}$
 & $5.36e^{9}$ 
 & $1.26e^{15}$ \\
 
Retrain
 & 57.43$_{\pm 0.2}$ 
 & 61.29$_{\pm 0.3}$ 
 & 58.60$_{\pm 1.0}$
 & 57.43$_{\pm 0.2}$ 
 & 61.29 $_{\pm 0.3}$
 & 58.60 $_{\pm 1.0}$
 & $5.36e^{9}$ 
 & $1.26e^{15}$ \\

MoDe
 & 50.92$_{\pm 0.5}$ 
 & 53.55$_{\pm 0.5}$ 
 & 56.38$_{\pm 1.0}$
 & 57.75$_{\pm 0.3}$ 
 & 60.00 (1.29$_{\pm 0.8}$) 
 & 61.30 (2.70$_{\pm 1.5}$)
 & $2.79e^{9}$ (1.92$\times$)
 & $6.57e^{14}$ (1.92$\times$) \\

FedAU
 & 14.1$_{\pm 0.9}$ 
 & 15.2$_{\pm 0.7}$
 & 14.3$_{\pm 1.6}$
 & 57.82$_{\pm 0.9}$ 
 & 60.95 (0.34$_{\pm 0.5}$) 
 & 60.12 (1.52$_{\pm 0.8}$)
 & $4.72e^{9}$ (1.14$\times$)
 & $1.11e^{15}$ (1.14$\times$) \\

NoT
 & 1.0$_{\pm 0.0}$ 
 & 1.2$_{\pm 0.2}$ 
 & 1.3$_{\pm 0.3}$
 & 57.51$_{\pm 0.5}$ 
 & 58.31 (2.98$_{\pm 0.4}$) 
 & 59.90 (1.30$_{\pm 0.3}$)
 & $6.43e^{9}$ (0.83$\times$)
 & $1.52e^{15}$ (0.83$\times$) \\

FedOSD
 & 53.10$_{\pm 0.6}$ 
 & 55.90$_{\pm 0.8}$ 
 & 51.33$_{\pm 1.1}$
 & 57.82$_{\pm 0.1}$ 
 & 60.81 (0.48$_{\pm 0.5}$) 
 & 60.80 (2.20$_{\pm 1.2}$)
 & $2.57e^{9}$ (2.09$\times$)
 & $6.07e^{14}$ (2.08$\times$) \\

FedQUIT
 & 53.18$_{\pm 0.5}$ 
 & 56.29$_{\pm 0.6}$ 
 & 49.10$_{\pm 1.0}$
 & 57.90$_{\pm 0.2}$ 
 & 61.53 (\textbf{0.24$_{\pm 0.4}$}) 
 & 57.90 (\textbf{0.70$_{\pm 0.9}$})
 & \textbf{$4.56e^{8}$ (11.75$\times$)}
 & \textbf{$1.07e^{14}$ (11.78$\times$)} \\

\bottomrule
\end{tabular}
}
\caption{\major{Performance (post--unlearning and post--recovery) on the Tiny-Shakespeare dataset for client unlearning. For each method, the table reports test, retain, and forget accuracies after the unlearning phase, as well as the corresponding metrics after the recovery phase. The Efficiency block summarizes the communication (bytes) and computational (FLOPs) costs incurred during unlearning and recovery. In parentheses, we also report the absolute gap with \textit{Retrain}, where lower values indicate better alignment with the ideal retraining baseline. Lower efficiency costs (higher relative gain in parenthesis) are preferable.}}
\label{table:lstm}
\end{table*}



\subsection{Ablation on Virtual Teacher's Non-True Structure}
\label{sec:ablation_teachers}
\begin{table*}[t]
\centering
\adjustbox{max width=\textwidth}{
\begin{tabular}{l l c c c c c c c}
\toprule
 &  & \multicolumn{3}{c}{\textbf{After Unlearning}} & \multirow{2}{*}{\textbf{R $\downarrow$}} & \multicolumn{3}{c}{\textbf{After Recovery}} \\
\cmidrule(lr){3-5}\cmidrule(lr){7-9}
 & \textbf{Method} & \textbf{Test Acc. ($\Delta$ $\downarrow$)} & \textbf{Retain Acc. ($\Delta$ $\downarrow$)} & \textbf{Forget Acc. ($\Delta$ $\uparrow$)} &  & \textbf{Test Acc. ($\Delta$ $\downarrow$)} & \textbf{Retain Acc. ($\Delta$ $\downarrow$)} & \textbf{Forget Acc. ($\Delta$ $\downarrow$)} \\
\midrule
\multirow{7}{*}{\rotatebox{90}{\textbf{IID}}}
 & Retrain             & 58.31 & 82.72 & 58.02 &   & 58.31 & 82.73 & 58.04 \\
 & \major{Random} & \major{1.00 (57.3)} & \major{1.00 (81.7)} & \major{1.00 (57.0)} & \major{16.2} & \major{58.50 (0.2)} & \major{74.50 (8.2)} & \major{69.00 (11.0)} \\
 & Incompetent         & 2.88 (55.4) & 2.88 (79.8) & 2.90 (55.1) & 15.7 & 58.81 (0.5) & 73.09 (9.6) & 67.61 (9.6) \\
 & Flatten             & 1.61 (56.7) & 1.61 (81.1) & 1.61 (56.4) & 16.9 & 59.20 (0.9) & 74.02 (8.7) & 67.10 (9.1) \\
 & Probability Ladder  & 13.61 (44.7) & 13.61 (69.1) & 13.05 (45.0) & 8.7 & 59.28 (1.0) & 74.90 (7.8) & 69.91 (11.9) \\
 & Logit Ladder        & 33.11 (25.2) & 37.24 (45.5) & 34.88 (23.1) & 10.8 & 59.25 (0.9) & 76.14 (6.6) & 66.71 (8.7) \\
 & TopK (K=5)          & 22.31 (36.0) & 25.16 (57.6) & 22.31 (35.7) & 12.1 & 59.26 (1.0) & 75.42 (7.3) & 61.31 (3.3) \\
 & FedQUIT             & 25.51 (32.8) & 27.23 (55.5) & 23.54 (34.5) & 10.0 & 58.70 (0.4) & 76.62 (6.1) & 59.42 (1.4) \\
\midrule
\multirow{7}{*}{\rotatebox{90}{\textbf{N-IID ($\alpha{=}0.1$)}}}
 & Retrain             & 51.02 & 64.61 & 33.52 &   & 51.02 & 64.61 & 33.52 \\
 & \major{Random} & \major{3.40 (47.6)} & \major{4.90 (59.7)} & \major{4.00 (29.5)} & \major{1.4} & \major{51.70 (0.7)} & \major{62.50 (2.1)} & \major{52.50 (19.0)} \\
 & Incompetent         & 18.30 (32.7) & 20.00 (44.6) & 12.90 (20.6) & 1.3 & 51.80 (0.8) & 62.30 (2.3) & 50.20 (16.7) \\
 & Flatten             & 16.02 (35.0) & 18.40 (46.2) & 2.10 (31.4) & 1.2 & 52.60 (1.6) & 63.80 (0.8) & 48.50 (15.0) \\
 & Probability Ladder  & 40.04 (11.0) & 47.42 (17.2) & 22.82 (10.7) & 1.3 & 52.00 (1.0) & 63.30 (1.3) & 47.20 (13.7) \\
 & Logit Ladder        & 46.93 (4.1) & 58.22 (6.4) & 18.64 (14.9) & 1.3 & 51.90 (0.9) & 63.90 (0.7) & 39.80 (6.3) \\
 & TopK (K=5)          & 42.74 (8.3) & 53.71 (10.9) & 7.30 (26.2) & 2.0 & 51.20 (0.2) & 63.40 (1.2) & 29.00 (4.5) \\
 & FedQUIT             & 45.24 (5.8) & 55.82 (8.8) & 8.02 (25.5) & 3.7 & 52.04 (1.0) & 63.22 (1.4) & 34.62 (3.5) \\

\bottomrule
\end{tabular}
}
\caption{Ablation on non-true structure of virtual teacher. Results just after unlearning and after recovery. $R$ means recovery rounds. In parenthesis, gap with \textit{Retrain}. ResNet-18, CIFAR-100.}
\label{table:ablation_teacher}
\end{table*}

\begin{table*}[t]
\centering
\adjustbox{max width=\textwidth}{ 
\begin{tabular}{l l c c c c c c c} 
\toprule 
& & \multicolumn{3}{c}{\major{\textbf{After Unlearning}}} & \multirow{2}{*}{\major{\textbf{R $\downarrow$}}} & \multicolumn{3}{c}{\major{\textbf{After Recovery}}} \\ 
\cmidrule(lr){3-5}\cmidrule(lr){7-9} 
& \major{\textbf{Method}} 
& \major{\textbf{Test Acc. ($\Delta$ $\downarrow$)}} 
& \major{\textbf{Retain Acc. ($\Delta$ $\downarrow$)}} 
& \major{\textbf{Forget Acc. ($\Delta$ $\uparrow$)}} 
& 
& \major{\textbf{Test Acc. ($\Delta$ $\downarrow$)}} 
& \major{\textbf{Retain Acc. ($\Delta$ $\downarrow$)}} 
& \major{\textbf{Forget Acc. ($\Delta$ $\downarrow$)}} \\ 
\midrule

\multirow{7}{*}{\rotatebox{90}{\major{\textbf{IID}}}}
& Retrain             & 58.31 & 82.72 & 58.02 &   & 58.31 & 82.73 & 58.04 \\
 & Flatten  ($\tau \rightarrow \infty$)           & 1.61 (56.7) & 1.61 (81.1) & 1.61 (56.4) & 16.9 & 59.20 (0.9) & 74.02 (8.7) & 67.10 (9.1) \\

 & \major{$\tau=4.0$}          
   & \major{1.80 (56.5)} & \major{1.90 (80.8)} & \major{1.80 (56.2)} 
   & \major{11.3} 
   & \major{58.74 (0.4)} & \major{74.92 (7.8)} & \major{68.36 (10.3)} \\

 & \major{$\tau=3.0$}          
   & \major{5.20 (53.1)} & \major{5.20 (77.5)} & \major{5.10 (52.9)} 
   & \major{10.0} 
   & \major{58.67 (0.4)} & \major{75.27 (7.5)} & \major{68.73 (10.7)} \\

 & \major{$\tau=2.0$}          
   & \major{14.65 (43.7)} & \major{15.57 (67.2)} & \major{13.56 (44.5)} 
   & \major{8.0} 
   & \major{58.53 (0.2)} & \major{74.54 (8.2)} & \major{65.47 (7.5)} \\

 & \major{$\tau=1.5$}          
   & \major{18.10 (40.2)} & \major{19.00 (63.7)} & \major{15.72 (42.3)} 
   & \major{9.7} 
   & \major{58.53 (0.2)} & \major{74.20 (8.5)} & \major{61.35 (3.3)} \\

 & FedQUIT ($\tau=$1.0)  & 25.51 (32.8) & 27.23 (55.5) & 23.54 (34.5) & 10.0 & 58.70 (0.4) & 76.62 (6.1) & 59.42 (1.4) \\

\midrule 

\multirow{7}{*}{\rotatebox{90}{\major{\textbf{N-IID ($\alpha=0.1$)}}}}
  & Retrain             & 51.02 & 64.61 & 33.52 &   & 51.02 & 64.61 & 33.52 \\
 & Flatten  ($\tau \rightarrow \infty$)   & 16.02 (35.0) & 18.40 (46.2) & 2.10 (31.4) & 1.2 & 52.60 (1.6) & 63.40 (1.2) & 48.50 (15.0) \\

 & \major{$\tau=4.0$}          
   & \major{22.90 (28.1)} & \major{27.90 (36.7)} & \major{0.90 (32.6)} 
   & \major{1.5} 
   & \major{51.98 (1.0)} & \major{63.20 (1.4)} & \major{46.07 (12.6)} \\

 & \major{$\tau=3.0$}          
   & \major{29.74 (21.3)} & \major{36.70 (27.9)} & \major{0.00 (33.5)} 
   & \major{2.0} 
   & \major{52.36 (1.3)} & \major{63.31 (1.3)} & \major{44.39 (10.9)} \\

 & \major{$\tau=2.0$}          
   & \major{35.85 (15.2)} & \major{45.30 (19.3)} & \major{1.40 (32.1)} 
   & \major{2.7} 
   & \major{52.21 (1.2)} & \major{63.60 (1.0)} & \major{42.25 (8.7)} \\

 & \major{$\tau=1.5$}          
   & \major{39.04 (12.0)} & \major{49.60 (15.0)} & \major{2.00 (31.5)} 
   & \major{2.7} 
   & \major{51.35 (0.3)} & \major{63.17 (1.4)} & \major{38.23 (4.7)} \\

& FedQUIT ($\tau=$1.0)  & 45.24 (5.8) & 55.82 (8.8) & 8.02 (25.5) & 3.7 & 52.04 (1.0) & 63.22 (1.4) & 34.62 (3.5) \\

\bottomrule 
\end{tabular} 
} 
\caption{\major{Effect of temperature scaling for virtual teacher's non-true structure ($\tau{=}1.0$ default choice for FedQUIT). Results just after unlearning and after recovery. $R$ means recovery rounds. In parenthesis, gap with \textit{Retrain}. ResNet-18, CIFAR-100.}} 
\label{table:ablation_temperature}
\end{table*}

\begin{figure*}[t!]
  \centering
\begin{subfigure}[t]{0.32\textwidth}
    \centering
    \includegraphics[width=\linewidth]{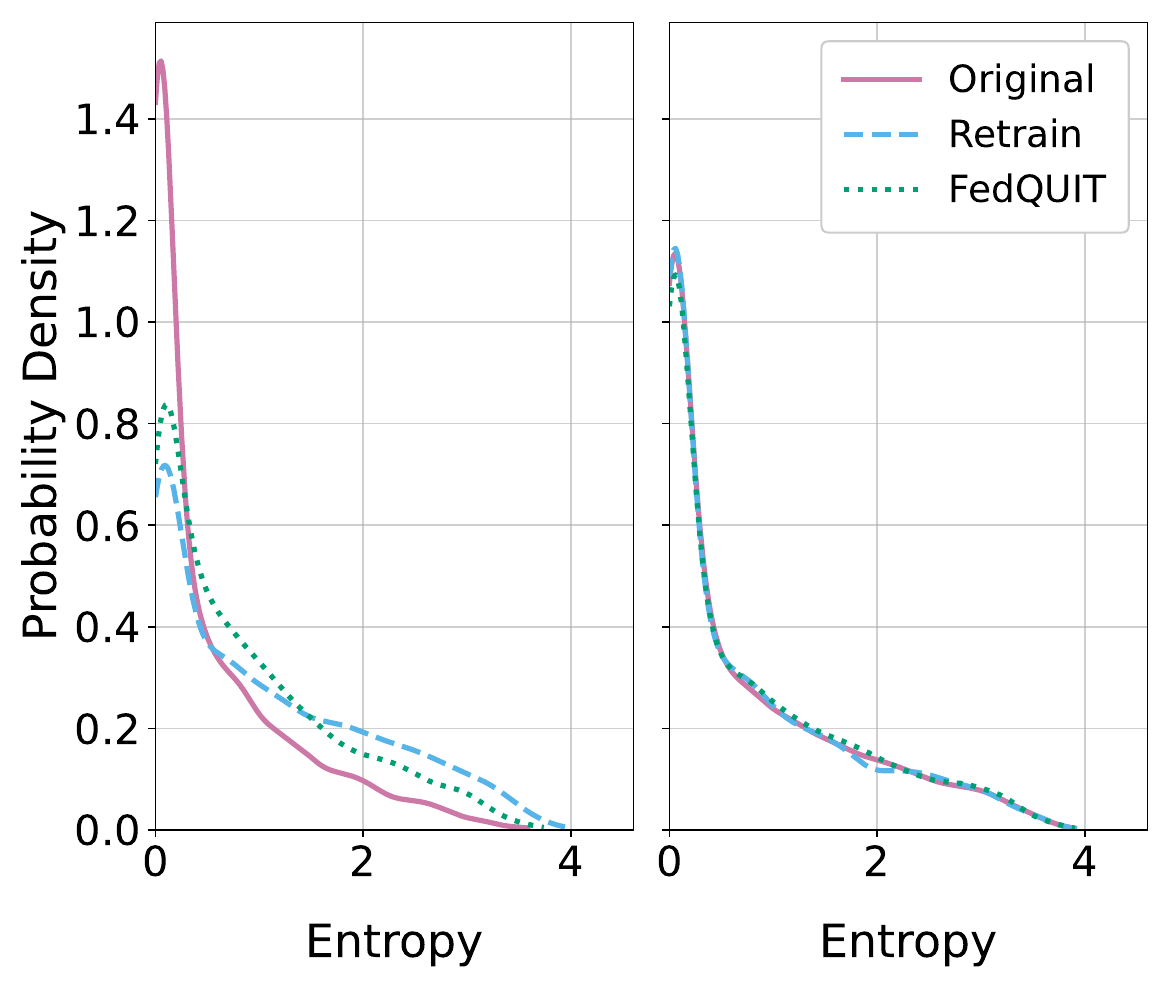}
    \subcaption{Entropy (test vs. forget).}
    \label{fig:sub_entropy}
\end{subfigure}
\hfill
\begin{subfigure}[t]{0.32\textwidth}
    \centering
    \includegraphics[width=\linewidth]{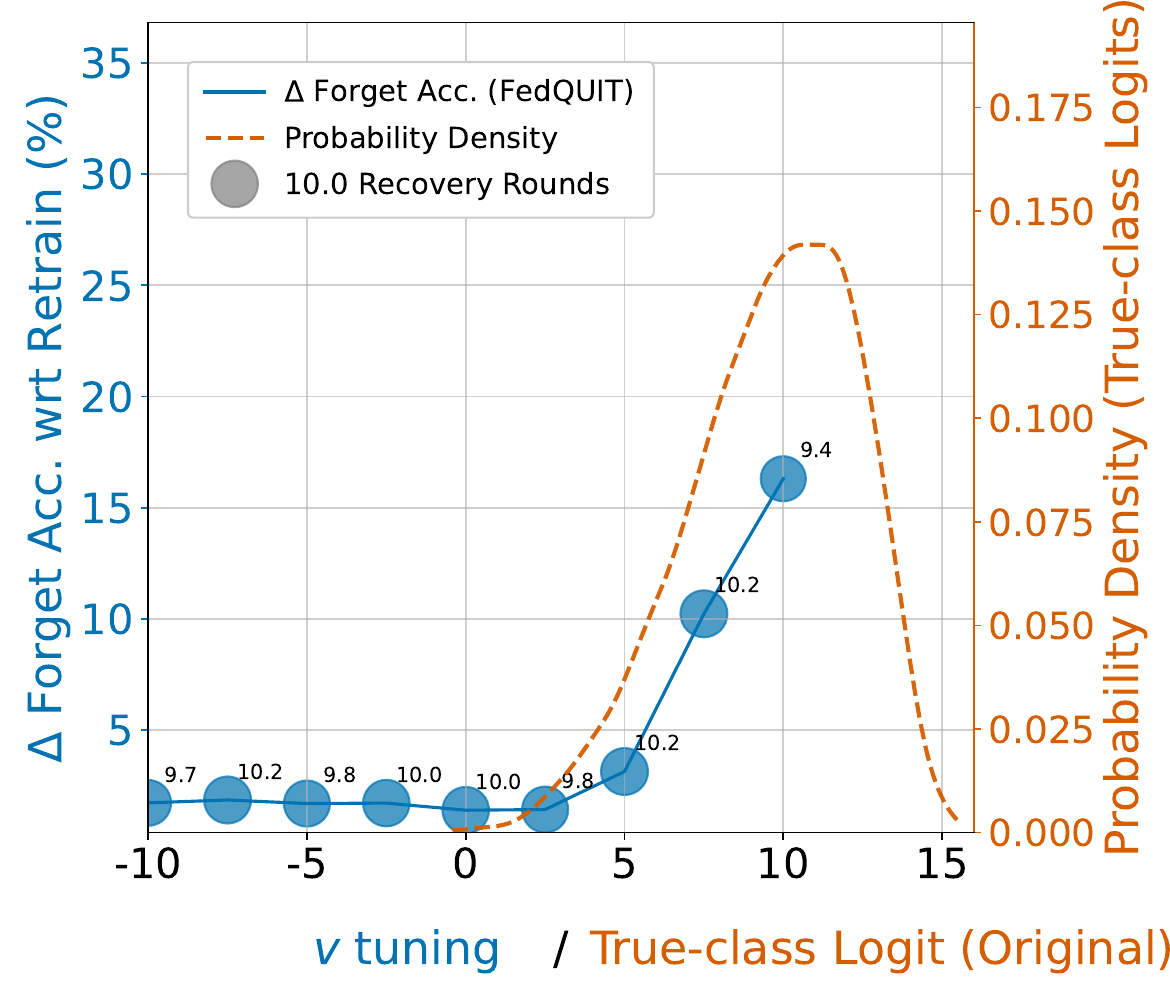}
    \subcaption{CIFAR-100, ResNet-18, IID.}
    \label{fig:sub_iid}

\end{subfigure}
\hfill
\begin{subfigure}[t]{0.32\textwidth}
    \centering
    \includegraphics[width=\linewidth]{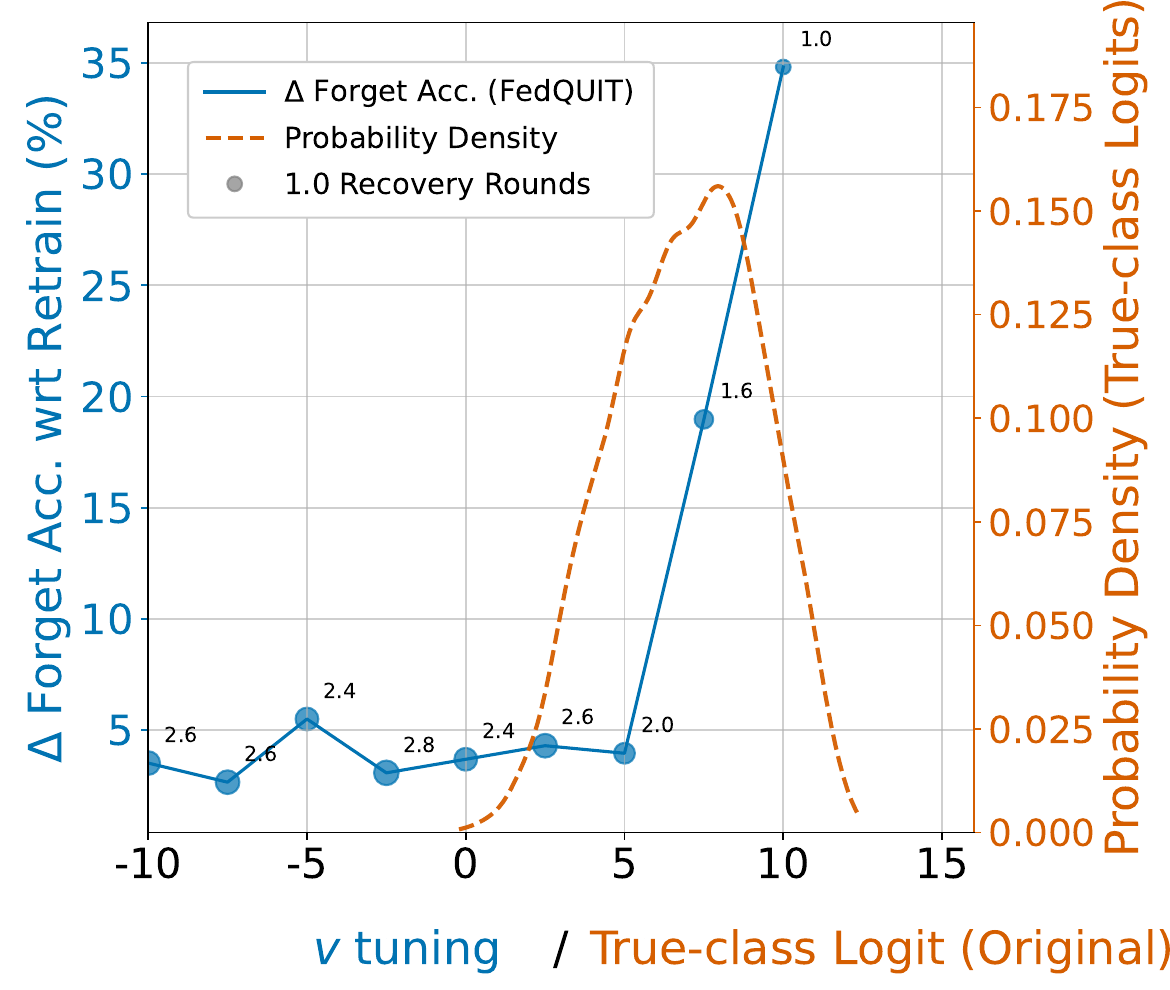}
    \subcaption{CIFAR-100, ResNet-18, N-IID.}
    \label{fig:sub_non_iid}
\end{subfigure}
\caption{
\textbf{(a)} Predictive-entropy densities of the global model on forget (left) and test (right) data. Setting: CIFAR-100 (Non-IID), MiT-B0.
\textbf{(b-c)} Forget efficacy of FedQUIT as $v$ is tuned (Eq.~\ref{eq:virtual_teacher_formula}).
Left $y$-axis: absolute after-recovery gap in forget accuracy ($\Delta$) w.r.t.\ \emph{Retrain} (lower is better). Right $y$-axis: probability density of the original model’s \emph{true-class logits} on the forget set (pre-FedQUIT). \emph{Shared $x$-axis:} the abscissa simultaneously denotes the tuning value $v$ (for the left-axis curve) and the true-class logit (for the right-axis density). Marker size encodes the number of recovery rounds.  
}
\end{figure*}

We test whether preserving the \emph{non-true} logit geometry during distillation is necessary to retain utility in FedQUIT. We construct a spectrum of virtual teachers that retain progressively more non-true structure: \emph{flatten} (uniform over non-true classes), \emph{top-K} (keeps only the head, collapses the tail), \emph{rank-only logit ladder} (preserves ordering, not gaps), \emph{probability ladder} (preserves ordering with equal probability steps), and \emph{FedQUIT} (preserves full non-true geometry). All variants keep the total exponential mass over non-true classes fixed, so the edited true-class probability is held constant; we also include an \emph{incompetent} teacher (fully uniform) and a \emph{random} teacher (random hard labels). Table~\ref{table:ablation_teacher} reports post-unlearning and post-recovery results. Complementarily, Table~\ref{table:ablation_temperature} increases the temperature applied \emph{only} to non-true logits, progressively smoothing the teacher and approaching the flattened case.
\textbf{Findings.} (i) \textit{Rank carries knowledge.} Immediately after unlearning, rank-preserving teachers retain higher test/retain utility than structure-free ones (logit ladder $>$ probability ladder $>$ flatten $\approx$ incompetent $>$ random), and under IID they often need fewer recovery rounds. (ii) \textit{Geometry drives recovery.} Matching \emph{Retrain} after recovery requires preserving gaps and tail mass (FedQUIT $>$ top-K $>$ logit ladder $>$ probability ladder $>$ flatten $>$ random), showing that full non-true geometry yields the best post-recovery behavior. (iii) \textit{Smoothing hurts.} Increasing non-true temperature monotonically degrades post-unlearning test/retain utility and shifts behavior toward flattened teachers, mirroring the structure ablations and weakening selective unlearning. Further analysis is in Appendix D, subsection J.


\subsection{Effect on Predictive Entropy}
\label{sec:predictive_entropy}
We analyze the Shannon entropy of softmax outputs in Fig.~\ref{fig:sub_entropy}, plotting the predictive-entropy density for \emph{Original}, \emph{Retrain}, and \emph{FedQUIT} on forget data (left) and test data (right). Predictive entropy provides a model-agnostic view of confidence: memorized training points typically elicit low-entropy, highly confident predictions, whereas removing the target data should increase uncertainty on that subset while leaving confidence on non-target data unchanged. \textbf{Findings.} On forget data, \emph{Original} is markedly overconfident (low entropy), consistent with having trained on those samples, whereas \emph{Retrain} shifts to higher entropy, reflecting increased uncertainty after removal. After recovery, \emph{FedQUIT} closely matches \emph{Retrain} on forget data (both shifted toward higher entropy) and is clearly separated from \emph{Original}, which remains concentrated at low entropy. On test data, the three distributions largely overlap, indicating that FedQUIT increases uncertainty primarily on the forget set while preserving behavior on held-out data.

\subsection{Sensitivity analysis on $v$}
\label{sec:sensitivity_on_v}


A key design choice in FedQUIT is the target true-class logit $v$ used to construct the virtual teacher: smaller $v$ should strengthen the forgetting signal, but overly aggressive values may be unnecessary. We therefore adopt the default $v_i=\min_{c} z^g_{i,c}$, which is parameter-free, data-adaptive, and theoretically motivated in Section \ref{sec:method}. Here, we validate it empirically by sweeping fixed $v$ values and measuring $\Delta$ Forget Accuracy w.r.t.\ \textit{Retrain} (lower is better) on IID and non-IID CIFAR–100 with ResNet–18. Results are reported in Figures~\ref{fig:sub_iid} and \ref{fig:sub_non_iid}.
\textbf{Findings.} As $v$ moves toward the left tail of that density, performance approaches \textit{Retrain} (stronger forgetting), in line with Proposition~\ref{prop:main-vmin}, which shows that decreasing $v$ increases the forgetting signal. We also compare two special choices in Appendix Table XX: $v$ to maximizes per-sample entropy ($v^\star$) and an extreme $v \ll \min_{c} z^g_{i,c}$ so that $p^{\mathrm{virt}}_{i,y_i}\approx 0$. The first underforgets (theoretical intuition in Appendix Proposition 3), whereas the latter is not significantly better than $v_i=\min_{c} z^g_{i,c}$. This supports the default parameter-free $v_i=\min_{c} z^g_{i,c}$.

\section{Conclusion}

We presented FedQUIT, a single-round framework for FU.
FedQUIT introduces a novel and general teacher–student formulation in which a
carefully crafted version of the global model’s outputs serves as a virtual
teacher. The framework naturally supports client-level and
sample-level unlearning, and extends to scenarios with multiple simultaneous
unlearning requests. We provided theoretical results showing that the proposed distillation-based
unlearning induces a bounded parameter shift and allows training to
resume with standard FedAvg convergence guarantees. This implies that
FedQUIT preserves the optimization properties of the original training process
while enabling efficient and controlled forgetting.

In our experimental evaluation, we compared FedQUIT with six state-of-the-art
baselines across four datasets, three model architectures, varying levels and
types of data heterogeneity, and different numbers of clients and participation
rates. The results show that FedQUIT achieves superior or comparable unlearning
efficacy while substantially improving efficiency in recovering retrain-level
model utility. To further strengthen the empirical study, we conducted extensive ablations on
the structure of the virtual teacher, the distillation temperature, the loss
function used during unlearning-oriented distillation, and both fixed and
data-adaptive penalization strategies. These analyses highlight the importance
of preserving non-true output geometry.

\bibliographystyle{IEEEtran}
\bibliography{bib} 
\vspace{-10pt}

\begin{IEEEbiography}[{\includegraphics[width=1in,height=1.25in,clip,keepaspectratio]{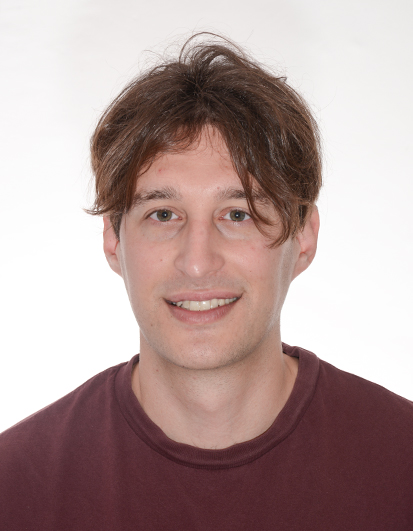}}]{Alessio Mora} received his Ph.D. in Computer Science and Engineering in 2023 from the University of Bologna, Bologna, Italy. He is currently an Assistant Professor at the University of Bologna. His research interests include decentralized learning, with a particular focus on federated learning and edge intelligence. He serves as an Associate Editor for IEEE Internet of Things Journal. He has served on the program committees of top-tier AI conferences, such as IJCAI and ICLR.
\end{IEEEbiography}

\vspace{-34pt}
\begin{IEEEbiography}[{\includegraphics[width=1in,height=1.25in,clip,keepaspectratio]{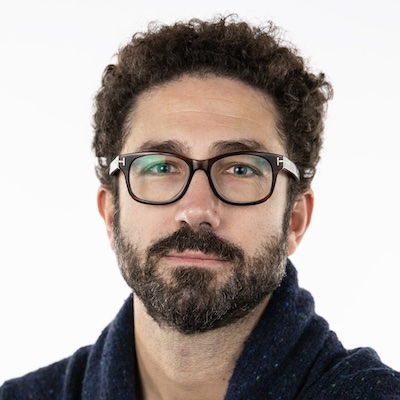}}]{Lorenzo Valerio}
is a senior researcher at IIT-CNR. 
%
%
His main research activity focuses on decentralized and resource-constrained machine learning, mainly targeting Edge/IoT environments.
%
%
He organized several international workshops, such as IEEE AOC’15 and IEEE PeRConAI Workshop (from 2021 to 2025). He guest edited for Elsevier Computer Communications, Elsevier Pervasive and Mobile Computing, and Springer Evolving Systems. He received three Best Paper Awards from international conferences and research institutes and one Best Paper Nomination. 
He is currently on the editorial board of the Elsevier Computer Communication journal and the PC of several conferences such as IEEE IJCNN and AAAI, IEEE MSN, and International Workshops.  
He is (or has been) active in several national projects under the NRRP/PON programs and international projects funded by the European Community under the HE/H2020/FP7 programs.
\end{IEEEbiography}

\vspace{-34pt}
\begin{IEEEbiography}[{\includegraphics[width=1in,height=1.25in,clip,keepaspectratio]{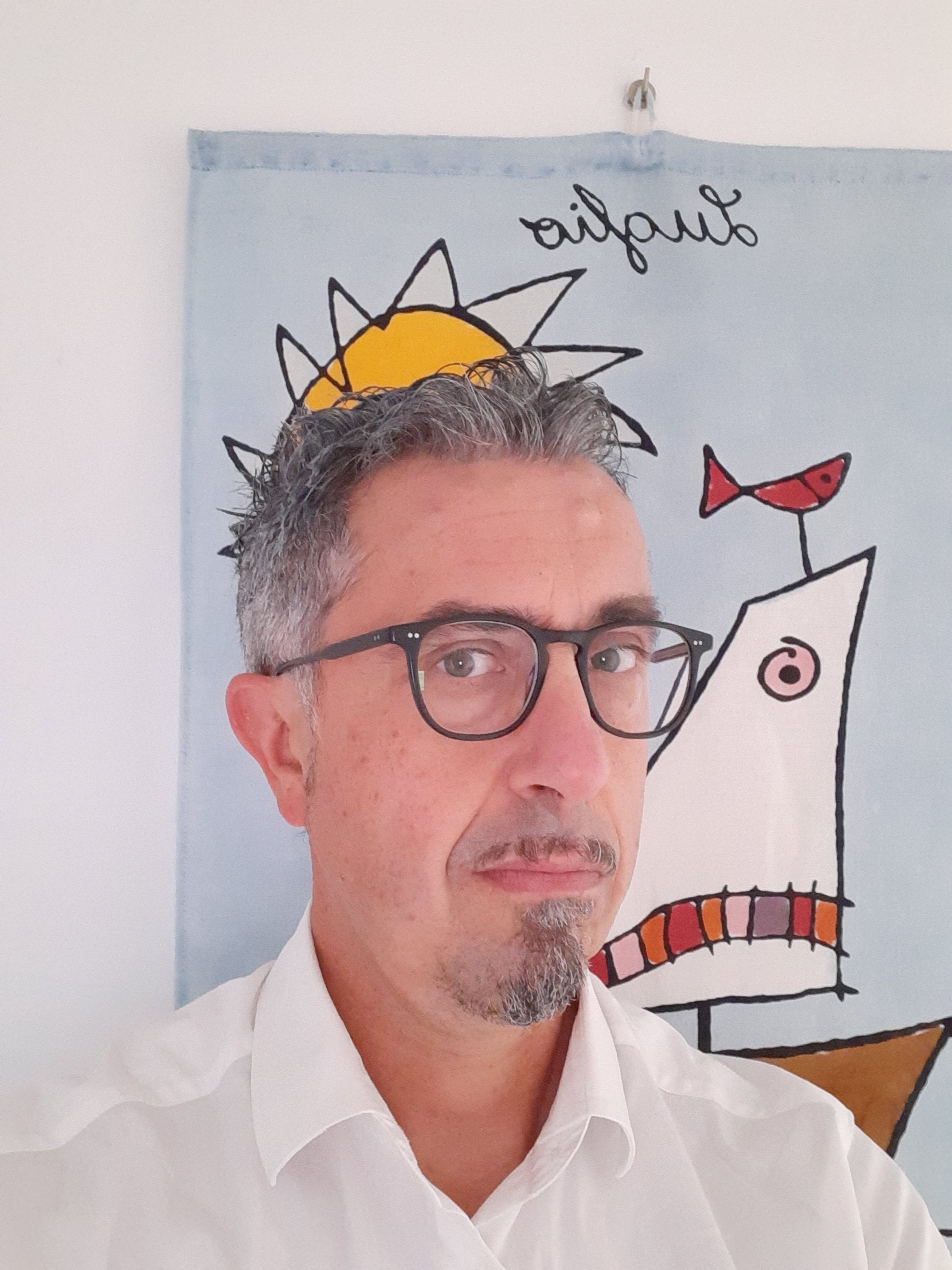}}]{Paolo Bellavista}
is professor of distributed and mobile systems at the University of Bologna, where he leads the Mobile Middleware research group (https://site.unibo.it/middleware/en). His research interests include middleware for mobile computing, dynamic QoS management in the cloud continuum, infrastructures for big data processing in industrial environments, digital twins for industrial automation and smart cities, and performance optimization in wide-scale and latency-sensitive deployment environments. He is Associate EiC of IEEE COMST and serves on the Editorial Boards of several IEEE/ACM/Elsevier international journals.
\end{IEEEbiography}

\vspace{-34pt}
\begin{IEEEbiography}[{\includegraphics[width=1in,height=1.25in,clip,keepaspectratio]{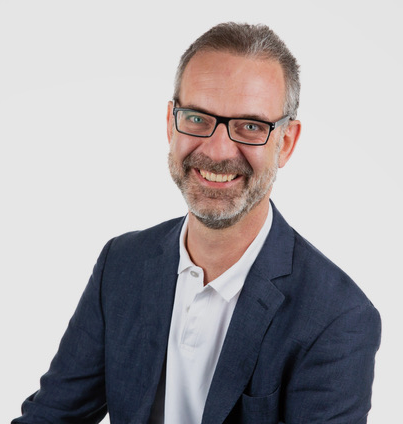}}]{Andrea Passarella}
(Ph.D. 2005) is a Research Director at the National Research Council of Italy (CNR) and Director of the Institute of Informatics and Telematics (IIT). He previously worked at the University of Cambridge Computer Laboratory (UK). He has authored 240+ publications on online and mobile social networks, decentralized AI, Next Generation Internet, and wireless networks, receiving several best paper awards. He is the founding Associate EiC of Elsevier Online Social Networks and Media journal.  He has chaired the IFIP WG 6.3 "Performance of Communication Systems" and contributed to several EU expert groups (FIRE, Networld Europe, NGI). He is currently the national coordinator of the ESFRI SLICES Research Infrastructure. He is/has been CNR (co-)PI in multiple projects on Human-centric AI, Big Data, Future Internet, and related areas.
\end{IEEEbiography}

\end{document}